# A trans-disciplinary review of deep learning research and its relevance for water resources scientists

Chaopeng Shen[1]

Civil and Environmental Engineering

Pennsylvania State University, University Park, PA16802

## Abstract

Deep learning (DL), a new-generation of artificial neural network research, has transformed industries, daily lives and various scientific disciplines in recent years. DL represents significant progress in the ability of neural networks to automatically engineer problem-relevant features and capture highly complex data distributions. I argue that DL can help address several major new and old challenges facing research in water sciences such as inter-disciplinarity, data discoverability, hydrologic scaling, equifinality, and needs for parameter regionalization. This review paper is intended to provide water resources scientists and hydrologists in particular with a simple technical overview, trans-disciplinary progress update, and a source of inspiration about the relevance of DL to water. The review reveals that various physical and geoscientific disciplines have utilized DL to address data challenges, improve efficiency, and gain scientific insights. DL is especially suited for information extraction from image-like data and sequential data. Techniques and experiences presented in other disciplines are of high relevance to water research. Meanwhile, less noticed is that DL may also serve as a scientific exploratory tool. A new area termed "AI neuroscience," where scientists interpret the decision process of deep networks and derive insights, has been born. This budding sub-discipline has demonstrated methods including correlation-based analysis, inversion of network-extracted features, reduced-order approximations by interpretable models, and attribution of network decisions to inputs. Moreover, DL can also use data to condition neurons that mimic problem-specific fundamental organizing units, thus revealing emergent behaviors of these units. Vast opportunities exist for DL to propel advances in water sciences.

Accepted for publication in Water Resources Research. Go to journal website for print-ready version

[1] Corresponding author: cshen@engr.psu.edu

## Table of Content

## 1. Motivations

Across various corners of the scientific world, scientists have come in contact with the term "deep learning" (DL). Deep learning consists of a collection of advanced Artificial Neural Network research which had gained momentum since 2012, when it started to become winners of machine learning contests. While some hype does exist, DL undeniably delivered unrivaled performance and solved exciting problems that have been difficult for artificial intelligence (AI) for many years (LeCun et al., 2015; Silver et al., 2016). DL algorithms have shown a generational leap in predictive capability which some argued as "unreasonable" (Baldassi et al., 2016; C. Sun et al., 2017). Since 2012, as an indication of advances, DL emerged as a dominant force that breaks records in most machine learning contests where it is applicable (Schmidhuber, 2015). While substantial manual effort has been spent on earlier methods such as Support Vector Machines and Hidden Markov Models, deep neural networks repeatedly show significant advantages over them and other statistical methods. DL models can digest large quantities of data and can often generalize successfully to new instances.

Powered by DL, AI is rapidly reshaping scientific research and our lives. On the scientific front, DL is transforming many scientific disciplines (Section 3) and is widely adopted and regarded as an indispensable tool of the future. By showing the features and ongoing success of DL across various disciplines, this paper demonstrates the potential of DL to propel water sciences. On the daily-life level, DL has gained wide adoption by nearly all industrial leaders in technology and is proliferating in a "Cambrian explosion" fashion (Leopold, 2017). The influence of DL-powered AI in the industry is well summarized by Google's co-founder Sergey Brin, who wrote in their annual Founders' Letter that "*The new spring in artificial intelligence is the most significant development in computing in my lifetime. Every month, there are stunning new applications and transformative new techniques*" (Brin, 2018). Object identification and image recognition technologies based on vision DL are being deployed in medical image diagnosis (Greenspan et al., 2016; Shin et al., 2016) and autonomous vehicles (C. Chen et al., 2015; Mariusz Bojarski et al., 2017). Machines can now understand human-voice instructions and human emotions thanks to speech recognition (Graves et al., 2013) powered by time series DL. Digital assistants, most powered by some forms of time series DL (Sak et al., 2015), will soon be able to have human-like situational conversations that are too natural to tell it is from a machine (Goode, 2018). Outside of information technology, AIs powered by DL are revolutionizing traditional industries like transportation, law, pharmaceutical, communication, manufacturing, healthcare, finance, accounting, nature resources, etc. The various industries would not be expected to choose DL models had these models not offered a significant boost in performance.

Although water sciences are not strangers to machine learning, the adoption of DL in water resources or hydrology has so far only been gradual (see summary in Section 3.1.3). Water sciences, and hydrology in particular, are now facing significant fresh and old challenges along with opportunities, for example:

- The interdisciplinarity and human dynamics challenge: water scientists' societal missions now mandate us to account for not only the movement and transformations of the water cycle, but also how it reciprocates and interacts with other subsystems such as biogeochemistry, climate, vegetation dynamics, biotic and abiotic soil environments, and human dynamics including socioeconomics (Sivapalan & Blöschl, 2017; Wood et al., 2011). Many of these systems were previously considered exogenous to hydrology, but as the close interconnectedness of these systems are uncovered (Cai et al., 2018; Russo & Lall, 2017), we need to consider them as endogenous system components. For example, we must now combat concepts and processes that may not have been formulated or are difficult to be, such as human responses to water shortages, evolving consumptive demands, stream health, or freshwater fish biodiversity. For another example, urban water infrastructure could be very complicated and time-consuming to model directly. Moreover, we have been called on, more than ever, to deliver accurate and socially relevant predictions, attributions, and optimizations in a timely manner.
- The data deluge and the data discoverability challenge: the diversity and volume of data that become available are both a blessing and a challenge. Satellites such as the Sentinel, Landsat, MODIS, soil moisture missions, and Gravity Recovery and Climate Experiment (GRACE) together log a large amount of multifaceted data. Emerging data sources such as miniaturized satellites, Interferometric Synthetic Aperture Radars, unmanned aerial vehicles, and camera rain gauges (Allamano et al., 2015), are presenting valuable data that were not available previously at such volumes (McCabe et al., 2017). So are monitoring networks such as pressure sensors in water infrastructure. These new sources of information are bound to enrich our understanding and empower our modeling capabilities. However, the data discoverability is an issue. Extracting comprehensible yet hidden information from the unprecedented volume of data or even devising approaches to extract such information can be daunting tasks. Conventional information retrieval approaches often require extensive human labor and domain expertise because they often require case-by-case calibration and adjustment. More importantly, without advanced techniques, we may not even realize what kind of hidden and abstract information that can be extracted nor recognize the limits of accuracy of such extraction, which results in under-utilization of available data. These practical aspects greatly limit their use in providing inputs and constraints to models.
- Unrecognized/un-represented linkages: models could produce errors if important processes and relationships are not included, due to either knowledge or capability limitations. For example, in the past decade the rudimentary hydrologic representations in Earth System models, especially lateral groundwater flow, have limited our capability to simulate land-atmospheric processes and multiscale feedbacks (Clark, Fan, et al., 2015). Inadequately represented processes such as preferential flows, ecosystem demographics (Moorcroft et al., 2001), rooting depth and root hydraulics (Fan et al., 2017) continue to affect model fidelity in relevant regions. In some cases, we have a crude idea of what processes/variables might be important, and have some data at hand, but implementing the corresponding model in an integrated modeling system might be too time-consuming, yet the relationships between these variables are complex and do not manifest themselves when they are simply plotted against each other. Some data-driven mechanisms to explore potential sophisticated linkages between variables would enable rapid hypothesis testing.

- The model scaling and equifinality challenges: These are long-standing modeling issues that have recently been shunned rather than addressed. It has long been known that process-based models' results can be scale-dependent, because governing equations such as Darcy's law and the Richards equation were applied at spatial scales incongruent with their laboratory or field derivations. When used in hydrologic models, they serve as phenomenological behavior laws with effective parameters, which adds to the scale dependence (Bloschl & Sivapalan, 1995; Troy et al., 2008; Wood, 1998; Wood et al., 2011). Fully resolving hydrologic dynamics at hyper-resolution (~$O(10^1$ m), which is needed for channel dynamics, is prohibitively expensive (Shen et al., 2016). How to effectively and efficiently resolve subgrid-scale processes and heterogeneity, e.g., in soil moisture and groundwater flow, remains a challenge in many disciplines including hydrology. Equifinality (competing models and parameter sets that can generate similar outcomes) (Beven, 2006) is another long-standing issue. Equifinality exists for both parameters and model structures. Although methods for testing multiple working hypotheses, e.g., Structure for Unifying Multiple Modeling Alternatives (SUMMA) (Clark, Nijssen, et al., 2015) are crucial approaches to tackling equifinality, they are nonetheless limited to those processes that have already been conceived. Again, it can be prohibitively expensive for humans to thoroughly explore the possible hypotheses.
- The need for regionalized parameters and models: with traditional data mining methods, data-driven hydrologic studies are often forced to regionalize or bin their datasets and create specialized models that fit those particular regions or bins. For example, the hydraulic geometry of channels have regionalized curves that apply poorly outside of the region from which they are built (NWMC, 2017; Vergara et al., 2016); effective hydrologic model parameters need to be regionalized (Beck et al., 2016; Troy et al., 2008) to produce good simulation results. This need also applies to the calibration of conceptual hydrologic models. Typical reasons for regionalization include a lack of knowledge of the underlying process or influential latent variable. A result of regionalization is that in each region the available data points are reduced and thus have less statistical power to condition the models.

In this paper I argue that DL can help address the abovementioned challenges: DL can serve as water scientists' operational approach to model interdisciplinary and difficult-to-model processes; It will be crucial for helping us handle the increasing volume and diversity of data, extracting useful and abstract information out of data; DL can help measure relevance of information for a given variable and present possible explanations; DL can help resolve scaling issues from a data-driven perspective and help build uniform parameterization schemes; DL can help capture high-dimensional, multi-modal data distributions and serve as realistic scenario generators; It may be possible to reveal fundamental organizing principles and emergent patterns. The new advances in DL interpretation present new opportunities to understand the network instead of using it as a blackbox. In Section 4, I describe why DL is relevant to these applications.

So, just what is deep learning and how is it different from previous machine learning methods? What type of problems is it suitable for and how can DL help in tackling some of the important challenges mentioned above? Can we use accurate DL models to advance our understanding? Because DL diverges significantly from water resources scientists' typical educational background, the community may benefit from clarifications of these concepts, a synthesis of the basic technical elements, as well as frontier DL research ideas.

The purpose of this review paper is not to give thorough technical descriptions or a historical recount of deep learning, which have been accomplished elsewhere, e.g., (Bengio et al., 2013; Hinton, Deng, et al., 2012; LeCun et al., 2015; Schmidhuber, 2015). Rather, I strive to provide a one-step launchpad for water scientists who are interested in exploring DL. This paper (1) provides a concise and simple technical overview of DL for water resources scientists (Section 2);(2) summarizes the progress in varied disciplines to provide a broad perspective of the state of deep learning research in sciences and progress relevant to water (Section 3); (3) review methods for interpreting DL networks (Section 3.2) and (4) examines potential directions where DL can contribute to solving challenges facing water science research (Section 4).

The interest in DL is rising among water scientists with diverse backgrounds. This review paper starts from a basic introduction to DL to its most recent applications in sciences and specifically, fields related to water. Therefore, readers are recommended to follow reading priorities as summarized in Table 1.

**Table 1. Recommended reading priorities for different audiences**

| For water researchers who… | Reading Priorities (Sections) |
|---|---|
| have an initial interest in DL but do not have much background in machine learning | 2 (with focus on 2.3.1- 2.3.3) & 4, A |
| want to simply use DL for extracting information from remote sensing images or time series data | 2.2, 2.3.2, 2.3.3 & 4 |
| want to use DL for capturing data distributions and generating scenarios | 2.2, 2.3, 2.5 & 4 |
| would like to know what DL has helped achieve in other scientific disciplines | 3.1 & 6 |
| are interested in using DL as a knowledge extraction tool or interpreting its results | 3.2 & 6 |

## 2. Deep learning basics

Here I attempt to give a self-contained technical background about DL, including some basic machine learning terminology (section 2.1), what is deep learning (section 2.2), some popular deep learning architectures (section 2.3), and hardware and software support (section 2.6). Readers who are familiar with machine learning concepts may skip section 2.1. For the purpose of comparison, I have also included brief descriptions of early-generation machine learning methods and their applications in hydrology in Appendix A.

### 2.1. Basic machine learning terminology

The machine learning literature is filled with an agonizing amount of jargons and terminologies. Unfortunately, knowing some of these terms is necessary for understanding the DL literature. Here I only introduce some basic ones relevant to the introduction of DL, while some other necessary ones are introduced in different sections. Those already familiar with machine learning should skip this section.

Machine learning can be broadly categorized into supervised learning and unsupervised learning. Supervised learning is trained to predict some observed target variables, either categorical or continuous, given some input attributes. The target variable is alternatively called dependent variables or labeled data. The algorithm could be configured for classification tasks, where we identify to which of a set of categories a new instance belongs, e.g., whether the image is a cat or a dog, or regression tasks, where a continuous prediction for the target variable is produced. Regression tasks are more often seen for scientific applications. An unsupervised learner is not given a target to predict. Unsupervised learning seeks to learn how to represent input data instances efficiently and meaningfully by finding hidden structures, e.g., principal components, clusters, and parameters of data distributions (Ghahramani, 2004). Principal component analysis, for example, seeks to find a few (uncorrelated) linear combinations that best represent the variations in the input data. Unsupervised learning is thus often employed for dimensional reduction (Redlich, 1993) or feature extraction (Coates et al., 2011; Ranzato et al., 2007; Sermanet et al., 2013), because after removing less important details, the remaining key bits of information are the most important to characterize the data.

Artificial Neural Networks (ANN) is a supervised learning approach to approximate functions by connecting units called neurons (Figure 1, often also called cells) across layers. A neuron on layer $l$ receives inputs from multiple neurons on layer $l$-1 and computes a linear or nonlinear transformation of the combined input, e.g., *sigmoidal*, *tanh*, rectified linear units (*ReLU*, essentially a 1:1 linear function with a minimum set to 0). Its output, called its activations, are then sent to layer *l+1*. The connections are represented by linear weights between neurons, e.g., $w_{ij}$ denoting the connection between the *i*-th neuron and the *j*-th neuron, which are on layer *l-1* and *l*, respectively. A multi-layer network as shown in Figure 2a is called "fully connected," as every neuron on one layer is connected to every neuron on an adjacent layer. The input and output layers have direct connections to inputs and outputs and thus are called visible layers, as opposed to hidden layers in the middle. Prior to the tuning of the network weights, the neural network is but an empty vessel. The network weights are found by minimizing a loss function, typically the distance between data and predictions. The tuning process is called training (Ratsch, 2004) and it is this process that defines the purpose of the network. The most common training method is backpropagation (Rumelhart et al., 1986), which is a form of gradient descend method that propagates the loss as a signal into the network and updates network weights based on the chain rule. Compared to other optimization methods such as evolutionary algorithms, backpropagation is highly efficient thanks to the differentiability and linear nature of the network connections. The number of neurons on a layer is called width while the number of layers is called the depth of the network. Hence, a "deep network" implies a large depth.

The testing error refers to errors incurred when the algorithm is applied to data not in the training set. The expected testing error for the background population is thus also referred to as the generalization error. After a machine learning algorithm is trained, it can be used to make inference, meaning forward runs that make a prediction when given new instances. When there are multiple prediction targets, e.g., simultaneously predicting groundwater level and soil moisture, the network is said to do multi-task learning. It is worth mentioning that while deep networks are also artificial neural networks, the term "ANN" now appears to be more often reserved for non-deep networks, and such meaning is used in this paper.

### 2.2. Deep vs. non-deep machine learning

Although there is no clear-cut definition for deep learning, it generally refers to large, multi-layer neural networks working directly on big, raw data. The most distinguishing characteristic of deep networks from non-deep ones is the depth of the network, provided by the stacking of multiple neuron layers in combinatorial and carefully designed architectures. The advantages of carefully designed depth have been described at length in (Schmidhuber, 2015) and are excellently summarized in (Goh, Hodas, et al., 2017): (1) it allows the automatic extraction and engineering of a cascade of abstract features from raw data (Bengio et al., 2013). These features (sometimes also called representations) are the discriminative information about the data. In contrast, with non-deep methods, useful input features must be constructed by humans with domain expertise, which is often the most labor-intensive step. Non-deep learning methods were mostly input-output template matching and in general do not learn advanced representations; (2) after the knowledge of how to extract these features is learned and stored in the form of trained network weights, these networks can enable “transfer learning” (Yosinski et al., 2014): reuse of the model trained on one task to another task. For example, image recognition networks trained on ordinary pictures can be transferred to remote sensing tasks and even gravitational wave observatory data (see Section 3.1). Transfer learning greatly extends the value of available training data; (3) increased network depth allows exponential growth of the ability to represent complex functions (Raghu et al., 2016). Given the same amount of neurons, networks with larger depth could have much stronger ability to better capture abstract spatial or temporal structures hidden in data (Eldan & Shamir, 2016), because information can take many more paths when neurons are stacked in a combinatorial fashion.

The scientific and practical significance of automatic extraction of features cannot be overstated. It is due to this feature engineering that machine learning models can effectively exploit the value of data in the absence of extensive domain expertise. It then follows that the automatic feature engineering should allow the algorithms to discover new concepts that are perhaps beyond what human experts have conceived. However, this is not to say that domain expertise is no longer needed. Domain expertise will be needed to frame questions, identify inputs, construct suitable model architectures and interpret results. There is also evidence that domain expertise of physical principles can improve machine learning outcomes (Ganguly et al., 2014; Karpatne et al., 2017).

To give a concrete example, Tao et al. (2016) compared retrieving precipitation from satellite images using an earlier-generation neural network system of theirs called PERSIANN-CCS (Hong et al., 2004) and a DL model (stacked denoising autoencoder, SDAE, see more details in Section 2.3.1). PERSIANN-CCS is a widely used operational system. Its inputs are a number of manually defined features, such as minimum/mean temperature of a cloud patch, cloud-patch area and shape index, mean and standard deviation of cloud-patch temperature, and gradient of cloud-top brightness temperature. These features were manually designed from knowledge of precipitation and experience. They were calculated from raw satellite data. With the SDAE model, they directly provided the raw satellite image itself without much pre-processing, and the training target was the difference between PERSIANN-CCS and the ground truth. The outcome was that the SDAE model removed 98% of the bias and reduced the variance by 31%. For another example, in the 2012 Merck chemical activity prediction challenge, a computer scientist team’s deep learning model won first place with no feature engineering and minimal pre-processing against models that required expert-defined feature (Joyce Noah-Vanhoucke, 2012). Their model also surpassed the Merck Corporation’s internal operational model, which was based on extensive expert-extracted features (Goh, Hodas, et al., 2017). This example again suggests that DL may automatically extract better features than experts. In the science application section (3.1), there are many more examples where DL and traditional models were contrasted.

Large-depth network structures provide the ability to model highly complex functions, spatiotemporal dependences, and data distributions from big data. For image recognition tasks, improvement could still be achieved with DL when the size of the training dataset was increased from 100M to 300M, while the performance of simpler methods would have stalled at a far lower data size (C. Sun et al., 2017). The ability to digest big data allows us to create a uniform model that assimilates all available data, utilizing commonality while identifying differences, as opposed to dividing data into multiple bins and modeling each bin separately. Previous remote sensing techniques mostly utilized multi-spectral information to extract features, which often need adjustments and calibration from image to image. Deep architectures such as convolutional neural networks (CNNs, to be described in Section 2.3) can now efficiently and automatically extract features that represent object shapes and textures. These features are often more powerful and robust clues than spectral signatures (Section 3.1.1).

If the power of DL simply arose from its depth, why has DL not materialized sooner? After all, earlier-generation multi-layer networks could also be easily configured with arbitrary depth. However, if we simply increase the number of layers in a "vanilla" earlier-generation multi-layer network, it becomes very difficult to train due to a key issue called "vanishing gradient" (Hochreiter et al., 2001) — the training signal becomes exponentially small as it propagates into the network, making backpropagation ineffective for deep networks. Similar issues exist with vanilla recurrent neural networks for sequence analysis. Advances in DL such as improved architectures, unsupervised pre-training, and weights sharing (to be explained in Section 2.3) resolved this issue and made the training more computationally efficient and more data efficient.

Contrasting DL to other earlier-generation machine learning algorithms (see their descriptions and applications in Appendix A), these earlier methods were mostly not designed with the relevant architectural elements to enable automatic extraction of features. Thus, their inputs features must be computed from raw data with formulas provided by domain experts. Many methods also do not scale well to the large datasets that DL can handle. These methods, like DL, face other issues of their own. For classification and regression tree (CART), the number of data points decreases exponentially at the lower-level of the classification tree, making it easy to overfit at these levels. Support Vector Machines (SVM) can engineer hyperplanes to separate instances, but it does not have the combinatorial depth. Therefore, SVM cannot extract abstract notions. Gaussian Process Regression (GPR) has elegant mathematical properties, works well for smaller datasets and provides uncertainty estimates, but it relies on local kernels. As a result, GPR has difficulty capturing highly complex, non-smooth functions, scaling to large dataset or generalizing to regions in the attribute space with few data points. For Genetic Programming (GP), while its explicit and transparent features are appealing, without artful controls the results can nonetheless be unintelligible. The overall idea of GP is very useful and will continue to be, but there is little evidence that GP can directly model big-data and highly-complex problems.

### 2.3. Popular deep learning network architectures

In this section, I introduce several popular deep network architectures. Two state-of-the-art, widely adopted architectures are CNNs for image tasks and long short-term memory (LSTM) for sequence learning tasks such as time series modeling, although CNNs are sometimes used in language modeling as well. Both architectures work directly on raw data. Before introducing their structures, I first discuss earlier-generation multi-layer networks and stacked autoencoders, as they are the preliminaries to deep networks. The deep belief network is introduced as a representative of Bayesian networks.

Figure 2a shows a "vanilla" multilayer perceptron (MLP) neural network with one hidden layer and two output variables. Although MLPs have been popular in hydrology (Maier et al., 2010), they are not the architecture of choice for modern machine learning tasks for several reasons: MLPs have a large number of free parameters and thus require very large datasets; they do not enforce rules that capture spatial dependences and local patterns; They also do not promote feature engineering or dimension reduction by the network, and thus face slow training issues.

#### *2.3.1. Autoencoders and stacked denoising autoencoders (SDAEs)*

Autoencoders (Ballard, 1987; Hinton & Salakhutdinov, 2006) are important structural elements for deep networks. They are called such because they are trained to reproduce their inputs, i.e., their success is measured by how close their outputs approximate their inputs after training. However, they have to do so while passing information through a layer with far fewer neurons, which forms an information bottleneck (Figure 2b). On the output side of the bottleneck, the network needs to reconstruct the inputs with condensed information: imagine $2^8$ bits of information comes in as inputs, only $2^6$ bits can pass through the bottleneck, but, based on these $2^6$ bits, the network needs to produce $2^8$ bits of outputs which are as close to the input as possible. As a result, the bottleneck layer must learn how to retain only crucial information pertaining to the structure of the data and discard non-essential details. The training process is really the conditioning of the network by the entire input dataset to perform dimensional reduction, somewhat similar to principal component analysis (PCA), but it can work much better than PCA (Hinton & Salakhutdinov, 2006). The hidden layers output abstract features engineered by the network. The trained weights are the optimized way of generating these features. The training of an autoencoder is said to be unsupervised pre-training because, during the process, no target data are provided. Autoencoders are fully connected so that if there are M input nodes and N hidden nodes, the weight matrix to be learned between them is of size M×N (plus bias parameters).

Denoising autoencoders (Vincent et al., 2008) extends from autoencoders. Instead of training the network to reproduce inputs exactly, we add noise to the target, e.g., by randomly setting some inputs to zero before training begins. This noise corruption, contrary to our intuition, helps to reduce overfitting (because noise serves to annihilate peculiarities or artifacts in the training dataset) and train more robust networks. More discussion related to this point is found in Section 2.4.

The stacked denoising autoencoder (SDAE), as the name suggests, stacks multiple denoising autoencoders layers so that the hidden representations from layer *k* are fed as inputs to layer *k+1* (Figure 2c). The width of the network shrinks as depth increases, ensuring higher-level, increasingly abstract concepts are captured at deeper levels. Each of the autoencoder units is trained to reproduce its inputs. They naturally form a cascade of layers that shrink in size going from input to output. The network is trained "greedily," one layer at a time (Bengio et al., 2007; Hinton et al., 2006), meaning that the first autoencoder is first trained to reproduce its inputs, and then we move to train the next autoencoder to reproduce its inputs, which are outputs from the first one. We do this until all autoencoder layers are trained. There is typically a fully connected output layer after all autoencoders in SDAE.

### *2.3.2. Convolutional Neural Network (CNN)*

A CNN is also composed of a cascade of layers that shrink in width from input to output (Figure 2e). To a hydrologic modeler or a numerical analyst, CNN is reminiscent of a geometric multigrid matrix solver. Like SDAE, the shrinking in CNN is to ensure the condensation of information into more abstract concepts at deeper layers. Each shrinking stage typically consists of one or multiple convolution layers and corresponding "pooling" layers (pooling suggests reductive operations) as explained below. First, a convolutional layer uses a local filter to convolve with its input. The output typically has the same dimension as input. Each output element results from the convolution between the input pixels in a small geometric neighborhood of this element and the filter; second, a pooling layer subsamples the output from the convolution. For example, a "max-pooling" layer with a stride of 2 for a 2D image will select the maximum value out of each $2 \times 2$ neighborhood and produce an output image a quarter of its input size. In contrast, the last output layer is typically a fully connected one.

CNNs and SDAEs look similar, but they have important differences. While SDAE layers are fully connected, the convolutional and pooling layers in CNNs only work on a local neighborhood, so a hidden unit is only connected to a small number of input neurons. As a result, a hidden unit has a "field of sight" that is smaller than the whole input layer (Figure 2e). This hierarchical design helps capture local geometric features, spatial patterns, and extract larger-scale representations in deeper layers. It also enables localization of input-output relationships, which are exploited in interpretive studies (more on this point in Section 3.2). On another note, it is important to note the learnable parameters for the convolutional layer are only those filters. The convolution is achieved by repeated applications of the same learnable filters, although they can be different for different layers. Because filters (one filter for each convolutional layer) have only a small number of elements and all neurons share it in a convolutional layer (called weight sharing), it greatly reduces the number of parameters compared to a fully connected layer. Therefore, CNNs are much easier to train than a vanilla, fully connected MLP.

*2.3.3. Long Short-Term Memory (LSTM)*

LSTM (Greff et al., 2015; Hochreiter & Schmidhuber, 1997) is a type of recurrent neural network (RNN) structure that learns directly from time series data. RNNs are called such because they work with sequential data and the output of each time step is fed as inputs to the next time step. LSTM is "deep-in-time" and can learn when to forget and how long to retain the state information. Previously, simple RNNs update only a single past state (Figure 2f). The cell state, $h^t$, serves as the memory of the system, somewhat similar to the state in auto-regressive models. $h^t$ is multiplied by learnable weights and then combined with inputs to evolve the state to the next time level. RNNs are trained using the backpropagation through time algorithm (Mozer, 1989; Werbos, 1990), which applies backpropagation after first unrolling the network. As with other backpropagation algorithms, the loss function is propagated backward to determine updates to weights. With simple recurrent networks, backpropagation encounters the vanishing gradient problem as discussed previously (Hochreiter, 1998; Hochreiter et al., 2001). Concisely, for RNNs, because states from earlier time steps have undergone multiplication by weights many times before it affects the output, the impacts of backpropagation on them become exponentially small. As a result, simple RNNs train extremely slowly.

The LSTM solution to this problem is the specially designed units called "*gates*" and *memory cells* (Figure 2f). These gates, which are themselves simply neurons with learnable weights, surround the cell memory ($S^t$) to control the flow of information. After training, the input gate controls what inputs are significant enough to remember. The forget gate decides how long and what past state memory should be retained. The output gate determines how much of the memory is used to produce the output. Together, they allow the network to remember information from the long past while discarding non-essential information.

*2.3.4. Deep Belief Network (DBN)*

DBN (Hinton et al., 2006) is a variant of multiple layer perceptron networks, and it also aims at reducing the dimensionality of input data. Similar to the way SDAEs are stacks of autoencoders, DBNs are stacks of Restricted Boltzmann Machines (RBMs), which learn how to reconstruct input distribution through a bottleneck layer (Figure 2d). An RBM consists of two layers of neurons, just like the input and hidden layers of an autoencoder. However, an RBM's connections are undirected (bi-directional). During training, the input layer (or visible layer) multiplies the inputs by its weights and applies the nonlinear transformations. During reconstruction, the activations are multiplied by the transposed version of the same weights and transformed again to obtain activations at the input layer. Then the training process adjusts the weights to minimize the differences between the statistical distributions of the reconstructed and input data. DBN is probabilistic in that it seeks to reconstruct the distribution of inputs rather than exactly the inputs themselves.

Similar to SDAEs, DBNs are trained "greedily" layer-by-layer. However, there is no requirement that upper-level layers must have fewer cells than lower ones. Moreover, DBNs are not recurrent networks and were not designed to specifically maintain time dependencies: the previous inputs and states of the network do not influence future predictions. Thus DBNs do not have system memory that is important for modeling time series.

### 2.4. Regularization (model complexity penalization) techniques

Simply put, "regularization" means adding a penalty to the network to reduce its complexity and prevent overfitting. In this section three regularization techniques--namely, norm regularization, early stopping, and dropout--are introduced. Overfitting occurs when a model so closely fits the available training data that it treats data noise or peculiarities as true data signals, resulting in reduced accuracy when making predictions about unseen instances. A frequent concern about deep learning is that deep networks must be prone to overfitting due to their enormous degrees of freedom. However, effective regularization techniques have been developed to penalize overfitting. Since the incipience of machine learning, regularization has always been a core and inherent component. Here I first briefly describe earlier regularization techniques and then introduce the newer inventions.

Given similar performances, conventional statistical wisdom would always prefer a more parsimonious model, which is reflected in model selection criteria such as the Akaike Information Criterion (AIC) (Akaike, 1974). For those unfamiliar with this concept, AIC and its variants allow one to evaluate the quality of alternative models while considering the number of parameters in each model:

$$AIC = 2k - 2ln(\hat{L}) \quad (1)$$

where $k$ is the number of parameters and $\hat{L}$ is the likelihood function of the model. This criterion penalizes overfitting and directs the choice to simpler models depending on the tradeoff between model complexity and performance. However, discarding predictors entails loss of information; that is, we sacrifice information for model robustness.

Instead of completely discarding information, norm-based regularization techniques apply penalization factors depending on how large the norms of the parameters are. For example, for a linear regression model $y = \boldsymbol{x}^T\boldsymbol{\beta}$, where $y$ is the predictand, $\boldsymbol{x}$ is a vector of predictors, and $\boldsymbol{\beta}$ are the linear coefficients (the bias parameter, $\beta_0$, is omitted), a regularized linear regression approach will seek to reduce the sum of the coefficients:

$$\boldsymbol{\beta} = \underset{\boldsymbol{\beta}}{argmin}\left(\frac{1}{2N}\sum_{i=1}^{N}(y_i^o - \boldsymbol{x}^{\boldsymbol{T}}\boldsymbol{\beta})^2 + \lambda\sum_{j=1}^{n}|\boldsymbol{\beta}|\right) \qquad (2)$$

where $y^o$ is observation, $N$ is the number of observations, $n$ is the number of coefficients, and $\lambda$ is a regularization parameter that determines how heavily the formulation penalizes large coefficients. The first summation sign contains the optimization problem without regularization, while the second term directs the choice to models whose parameters are of smaller magnitude, imposing the law of parsimony. With a pre-defined $\lambda$, the algorithm holistically assigns less weight to less important parameters. However, it retains all the predictors instead of completely discarding insignificant ones. Real-world problems are often highly complex, with myriad factors each contributing a small influence on the outcome. Assigning proper weights to them appears to be a more appropriate treatment.

Norm-based regularization has been widely employed in early-generation machine learning including ANN. Another commonly used regularization is early stopping (Yao et al., 2007), where the training of a learner is stopped before the learner fully achieves its best performance. Early stopping criteria are often based on a separate validation dataset and some smoothness criteria, e.g., it stops when the error on the validation dataset starts to increase compared to earlier epochs (Prechelt, 2012).

The abovementioned techniques are applicable to deep networks, but DL has several new techniques that are especially helpful, including Dropout (Hinton, Srivastava, et al., 2012; Srivastava et al., 2014), without which DL could not have made much progress. Dropout randomly blocks a fraction (called the dropout rate) of the connections in a network, as illustrated in Figure 3. This operation is implemented via a randomly generated mask with the same dimension as the connections between two layers. This mask is 0 for the blocked connections and 1 for those kept open. One may expect this procedure to introduce noise into the optimization process, and it does, but just like the case with SDAE, adding noise, in fact, improves robustness. There are several interpretations of dropout: (1) every new dropout mask effectively creates a new, reduced-order subnetwork (Figure 3). These subnetworks then form an ensemble of small networks. Making inference using a trained deep network is akin to using the ensemble mean to make predictions, which is more robust (Baldi & Sadowski, 2014; Hara et al., 2016); and (2) because connections are randomly blocked, neuron weights cannot adjust at the same time to cancel each other's effects to fit the target (Hinton, Srivastava, et al., 2012). The simultaneous adjustment, termed coadaptation, is a primary reason for overfitting. For other interpretations, see Gal & Ghahramani (2015, 2016).

In practice, dropout and its variants have been found to be miraculously effective (Smirnov et al., 2014). However, standard dropout schemes that work for CNN do not work for recurrent networks like LSTM. If dropout is blindly applied to the recurrent connections (between $S^t$ and other cells in Figure 2f) in the network, the recurrence amplifies the introduced noise, leading to loss of memory, which hinders LSTM from encoding long-term information. Therefore, *Zaremba et al.* (2015) proposed to avoid dropout for recurrent links and apply dropout only to forward connections, i.e., the connections between $x$ and all four gates, and between output node and target. *Gal and Ghahramani* (2016) proposed to also dropout recurrent links, but the dropout masks must not change across time steps. *Semeniuta et al.*, (2016) further applied dropout to the recurrent link from the input gate to the cell memory, also with a constant-in-time mask. *Fang et al.* (2017) (hereafter termed FSKY17) applied the dropout of both *Gal and Ghahramani* (2016) and *Semeniuta et al.* (2016). In preliminary studies for FSKY17, if dropout were not applied, the network would achieve very small training error, but the testing error would be very large.

### 2.5. Generative Adversarial Networks (GAN)

GAN was a progress milestone in deep learning (Goodfellow et al., 2014; Mogren, 2016) motivated by the need to model high-dimensional, multi-modal distributions. GAN is "generative" in that it can generate, or, quite vividly, "imagine" new instances that resemble the training data sometimes to such a remarkable degree that humans can hardly tell they are not real. Powered by DL's ability to automatically extract features, GANs can capture data characteristics which are perceivable by humans but difficult to qualitatively describe, such as art styles. The generation process is equivalent to implicitly sampling from the modeled distributions. When we employ a loss function such as mean sum of squares, we implicitly make assumptions that the residuals are from independent Gaussian distributions. Such assumptions work poorly for multi-modal data, and Gaussian loss often guides the model to generate instances that appear too smooth. Importantly, GANs avoid directly making such assumptions, and are able to generate instances that respect the spatiotemporal dependencies in data. Before GAN, belief networks, variational autoencoders (Kingma & Welling, 2013), Markov Chain Monte Carlo (Brooks et al., 2011), and Boltzmann machines were all designed to model the data distributions in different manners, but they each suffer from computational inefficiency or complexity issues. The family of models that require explicit, tractable density functions is limited, while Markov Chain methods are very computationally expensive and their convergence can be very slow and unverifiable for high-dimensional space (Goodfellow, 2016).

A few years after its introduction, GAN has become a major driver for advances and is opening up a sub-discipline in AI research. GANs have been trained to generate images described by an input sentence (Sriram et al., 2017); GANs can transfer Monet's or painting styles to new raw pictures (J.-Y. Zhu et al., 2017); GANs can impressively evolve the age of persons on an image (Antipov et al., 2017), predict the next framework of videos (Mathieu et al., 2015), or produce super-resolution images from coarse ones (Ledig et al., 2016); they have also helped to train deep networks (Ororbia et al., 2016) more robustly.

The most attractive feature of GAN is that it can capture complex, multi-modal and conditional distributions without having to make explicit assumptions about the distributions. However, training GANs can be tricky and involves delicate procedures (Goodfellow, 2016).

In Geosciences and water sciences, GAN applications also start to emerge in a probabilistic inversion of geologic media and urbanization projections (to be reviewed in Sections 3.1.2 and 3.1.3).

The main idea of GAN is to simultaneously train two deep networks that are playing a zero-sum game against each other. The first network generates new instances and the second network is trained to find clues to detect whether its input comes from real data or the output of the first network. Thus these two networks are placed in competition – the first network tries to fool the second with counterfeit data and the second, with access to true data, tries to avoid being fooled. Another way to interpret it is that the second network tries to find artifacts produced by the first. When the second succeeds, it provides this feedback to the first, helping it improve.

Some equations here can be helpful in linking the concepts to the procedures. In the initial version of GAN, let $D(x)$ denote the detector-predicted probability that an instance $\boldsymbol{x}$ comes from the true data generating distribution, $p_{data}(\boldsymbol{x})$, while $G(\boldsymbol{z})$ denotes an instance produced by the generator given random noise $\boldsymbol{z}$ from a noise vector space $p_{\boldsymbol{z}}(\boldsymbol{z})$, the following value function (a cross-entropy function) is defined:

$$V\left(\theta^{(D)}, \theta^{(G)}\right) = \mathbb{E}_{\boldsymbol{x}\sim p_{data}(\boldsymbol{x})}[\log(D(x))] + \mathbb{E}_{\boldsymbol{z}\sim p_{\boldsymbol{z}}(\boldsymbol{z})}\left[\log\left(1 - D\left(G(\boldsymbol{z})\right)\right)\right] \tag{3}$$

where $\theta^{(D)}$ and $\theta^{(G)}$ are the network parameters of the detector and generator, respectively. The value of the first expectation will be the highest when all instances from true data is judged as true by the detector, while the second will be when all counterfeit instances are judged as false. The two networks play a minimax game:

$$\min_{\theta^{(G)}}\max_{\theta^{(D)}} V\left(\theta^{(D)}, \theta^{(G)}\right) \tag{4}$$

which, to put simply, means that the generator attempts to minimize $V$ (maximally confuse the detector) without knowing the detector's actions (without access to real data), while the detector attempts to maximize $V$ (make the most correct decisions), knowing the generator's actions (having access to ground truths to evaluate $V$). In the implementation, $-V$ is the loss function for the detector while $V$ (or a non-saturating loss, *i.e.*, $-\mathbb{E}_{\boldsymbol{z}\sim p_{\boldsymbol{z}}(\boldsymbol{z})}\left[\log\left(D\left(G(\boldsymbol{z})\right)\right)\right]$) is the loss function for the generator. In practice, one network is trained a few iterations while holding the parameters of the other network constant, before rotating to train the other network. Ideally, at equilibrium, the detector is no longer able to find reliable clues to discern between real and generator-produced instances. Then, the generator has successfully learned the data-generating distributions and the dependence structures of data.

### 2.6. Hardware innovations and software support

The advances in DL are inseparable from hardware innovations. Without drastic advances in hardware and, more importantly, the recognition of the use of more efficient parallel-computation hardware like Graphical Processing Unit (GPU) for the task, DL would not have been plausible (Schmidhuber, 2015). Although some concepts of deep learning were decades old, their popularity did not grow until 2010, when GPUs were found to be suitable for the task. GPUs now offer thousands of times more matrix computing power than central processing units (CPUs), and their growth continues (Greengard, 2016) or even surpasses (Hemsoth, 2016) Moore's Law (loosely, the observation that the number of transistors per square inch on integrated circuits had doubled every two years since their invention). This trend suggests there is significant room for growth for larger and more complex networks. However, the computing hardware landscape for DL is evolving rapidly and dynamically. At present, most training tasks take place on GPUs or a few AI-specialized hardware, while inferencing tasks may occur on CPUs, GPUs, or field-programmable gate arrays (FPGAs). More recently, as some believed that the computing requirements appear to have stabilized, application-specific integrated circuits (ASICs), which are chips specifically designed for some narrow computing tasks, start to emerge for both training and inference DL applications.

A number of open source software platforms are available for building DL models. Like hardware, the software landscape is also evolving rapidly. At the time of this writing, Tensorflow is a popular Google-supported DL package based on the Python programming language. Torch is based on a scripting language called Lua, but it also has a Python version called PyTorch which has enhanced functionalities. Both Tensorflow and PyTorch provide access to low-level controls of operators and loss functions. Tensorflow has a longer history and a larger user community. PyTorch appears easier to learn and experiment with. It also enjoys strong optimization support from the fast GPU deep learning library cudnn. Other libraries include Theano, Caffe, MXNet, CNTK, etc. Keras is also a popular framework that serves as an easier front-end to Tensorflow and CNTK, but access to modifying low-level details can be tricky. Since the number of DL frameworks is large, there are efforts such as the Open Neural Network Exchange that attempts to make networks created under different frameworks transferrable to each other.

### 2.7. Understanding the generalization power of DL

Although it was proven decades ago that neural networks can approximate almost any function (Hornik et al., 1989), many are intrigued by why deep networks generalize so well despite their massive size. Most attribute DL's power to big data, regularization techniques and advanced hierarchical architectures. Another mainstream viewpoint is that DL does not overfit because we use a two-step procedure, where a test set is used to constrain the resulting models (Kawaguchi et al., 2017). However, there are also significant debates. *Zhang et al.* (2016a) showed empirically that deep network can easily memorize details and even noise, and questioned if previous interpretation (regularization) was the true cause of generalization. *Arpit et al.* (2017) presented a rebuttal where they showed deep networks tend to prioritize learning simple patterns first and do not just memorize data. Regularization can degrade network performance on training data and hinder memorization without compromising generalization on test data. DL has also be shown to be related to renormalization group, an important theoretical physics technique which also extracts features at different length scales (Mehta & Schwab, 2014).

## 3. Trans-disciplinary applications of DL and its interpretation

### 3.1. DL applications in sciences

DL is gaining adoption in a wide range of scientific disciplines and, in some areas, has started to substantially transform those disciplines. Several physical science disciplines with big data sources have embraced DL, customized network structures for their needs, and apparently built community software packages. Here the purpose of reviewing these studies is that there could be a latecomer advantage for water sciences if we learn from the experiences of other domains: we can see what kinds of problems DL is well suited for and the state at which DL is adopted in various disciplines. We can also observe how other areas have customized and utilized DL. I introduce these applications only with a level of details relevant to water scientists and emphases are put on the main messages we can learn. Section 3.2.4 reviews the interpretive DL studies in sciences that focused on gaining insights from trained networks.

#### *3.1.1. In physical sciences*

In high-energy physics (HEP), DL was used to handle the large volumes of data produced by particle colliders for the detection of abnormal particles. Transforming their problems into image recognition problems, scientists found that DL was able to increase the statistical power of data. Previously, using shallow machine learning, the progress was slow as manual extraction of features was involved. Researchers showed that deep learning can learn more complex functions that enabled better discrimination between signal and noise (Baldi et al., 2014). The authors trained autoencoders to predict the existence of abnormal particles based on low-level, non-image inputs. They showed that the deep network did not require manual extraction of high-level features, yet it attained 8% improvement over earlier approaches. Where current analysis techniques lack the statistical power to cross the traditional significance barrier, the use of deep networks enhanced statistical power and is equivalent to a 25% increase in the experimental dataset (Baldi et al., 2015). Further, organizing collider data as images (called jet images), scientists found CNNs to be successful in identifying diverse particles (Aurisano et al., 2016; Butter et al., 2017; Komiske et al., 2017; de Oliveira et al., 2016; Racah et al., 2016) and classifying physical events (Racah et al., 2016).

Since telescope data come in great volumes, astronomy is a prime example of how DL helps address the data deluge. DL significantly improved data discoverability and reduced labor cost in extracting abstract metric information. For example, CNNs allowed automatic extraction of galactic model parameters from telescope images (Hezaveh et al., 2017). CNNs can work at an accuracy similar to complex approaches but millions of times faster. *George et al.* (2017) obtained promising results with CNN pre-trained by real-world images to detect artifacts in data collected in Laser Interferometer Gravitational-wave Observatory (LIGO). This use of pre-trained network is the transfer learning concept explained earlier. *Tuccillo et al.* (2016) demonstrated the potential of using CNNs to estimate galaxy morphologic parameters. Although mixed results have been observed in some cases, many are optimistic that DL will become the method of choice for data processing problems in astrophysics and other fields.

With large databases of genomes and complicated relationships that are not fully understood, the field of computational biology has demonstrated how DL may directly learn from data and offer new knowledge (Min et al., 2016; Park & Kellis, 2015). DL can learn directly from genomic sequences to predict which genetic sequences exhibit interactions with certain proteins (Alipanahi et al., 2015). Having built a domain-specific DL package they call DeepBind, the authors further perturbed the trained models to examine how genetic mutations may potentially lead to diseases: their models predicted known genetic variations that cause tumor risks, as well as potentially damaging mutations that were previously not known. DL can capture nonlinear dependencies that could help span multiple genomic scales, potentially generalizing beyond available data (Angermueller et al., 2016). Stacked autoencoders, DBNs, and LSTM could predict protein structures directly from their sequences with much better accuracy than conventional methods (Heffernan et al., 2015; Lyons et al., 2014; Sønderby et al., 2015; Spencer et al., 2015).

Three trends emerge from computational chemistry literature: (1) DL models working directly with (more) raw-level data seemed to perform on par with or better than those working with predefined features or descriptors (Goh, Siegel, et al., 2017; Gomes et al., 2017; Lusci et al., 2013; Mayr et al., 2016; S. L. Yang et al., 2014); (2) multi-task learning may produce stronger models as they develop shared feature extraction pipelines and imposing stronger constraints (Goh, Hodas, et al., 2017; Ramsundar et al., 2015; Schütt et al., 2018); (3) scientists could customize DL to emulate atoms and molecules in order to learn emergent relationships between structures and resulting properties. *Lusci et al.* (2013) used a novel graph structure (Koller & Friedman, 2009) to mimic molecular configurations and predict molecular properties. Other authors extracted CNN's hidden layer representation as chemical fingerprints to predict different properties (Coley et al., 2017; Schütt et al., 2017) (more discussion in Section 3.2.4). Moreover, the community has assembled large datasets and standard benchmarks (Wu et al., 2017) and a suite of DL packages like deepChem (Subramanian et al., 2016).

*3.1.2. In Geosciences: Remote Sensing, Global Change, and Climate*

In fields close to water such as remote sensing, DL is growing to be the preferred method of choice, proving it a crucial tool in discovering information from raw images (L. Zhang et al., 2016). Since CNNs excel in extracting information from geometric shapes, textures, and spatial patterns, they easily outperform earlier methods that only utilize spectral signatures or handcrafted features (Makantasis et al., 2015). Main RS applications (X. X. Zhu et al., 2017) include using CNN to classify or segment images (assigning classes to each pixel on an image for what they are, e.g., land use classes, crops types) (Geng et al., 2015), object recognition (finding targets from a series of images) (Wagner, 2016), object localization (Long et al., 2017), and terrain attribute extraction, e.g., sea ice concentration (L. Wang et al., 2016). Remote sensing applications often started with standard CNN and RNN designs, and then made incremental modifications for efficiency and accuracy gains (Long et al., 2017; Mou et al., 2017; X. X. Zhu et al., 2017). In global change analysis, DL models show advantages in estimating crop yield (Kuwata & Shibasaki, 2015; You et al., 2017). *Pryzant et al.* (2017) forgo the conventional spectral features method in favor of a combined CNN-LSTM model to estimate outbreaks of wheat fungus in Ethiopia. Their LSTM is stacked on CNN-extracted feature representations to incorporate both spatial structural and temporal change. In addition, CNN is used for overall interpretation of images such as scene classification (for an image, recognize a theme from a list of possible themes) (Marmanis et al., 2016; Nogueira et al., 2017), change detection (Puzhao Zhang et al., 2016), and object detection, e.g., vehicles (X. Chen et al., 2014). Such high-level tasks were difficult to achieve using earlier machine learning techniques.

In disaster detection and categorization studies, researchers have started to employ DL to detect wildfires (Lee et al., 2017; Sharma et al., 2017; Q. Zhang et al., 2016) and landslides from remote sensing images (Ying Liu & Wu, 2016). *Liu and Wu* (2016) applied pre-processing steps including discrete wavelet transformation and noise corruption and trained an SDAE to identify landslides on the transformed image. They argued that the transformation is necessary because the resolution of remote sensing images is too low. However, they did not directly show results to support the claim.

Furthermore, the remote-sensing community has exploited transfer learning and data augmentation and has adapted available architectures to suit their data quantity. For example, publicly available trained networks such as GoogLeNet (Szegedy et al., 2015) can be transferred for scene classification for satellite images (Hu et al., 2015; Marmanis et al., 2016; Nogueira et al., 2017). Transfer learning works because the hidden layers that have been trained to distill shape information are also effective even when ported to remotely sensed scenes. Transfer learning also means that certain expertise obtained from training on existing datasets can be modularized, packaged, and assembled. Data augmentation (Ding et al., 2016; Morgan, 2015) means increasing the training data by making perturbations to data that should not have mattered, such as rotation, translation, interpolation, elastic distortions, and affine transformations, etc.

In climate science, the number of applications of deep learning in climate modeling starts to rise quickly, with applications focusing on (1) identification of extreme climate events and (2) addressing the resolution challenge. In a study carried out at the Lawrence Berkeley National Lab, *Liu et al.* (2016) trained a CNN with two convolutional layers to detect extreme events using thousands of images of tropical cyclones, weather fronts, and atmospheric rivers. This new system achieves 89%-99% accuracy in detecting extreme events and is useful for benchmarking climate models. A significant amount of attention has been paid to using deep learning for precipitation forecasting, (e.g., Hernández et al., 2016; Shi et al., 2017; Pengcheng Zhang et al., 2017).

The model resolution has been a central challenge for climate modeling. For dynamic modeling, researchers trained dynamic convolutional layers, i.e., filters with weights that are dynamically updated using inputs during forward runs, in short-range weather predictions (Klein et al., 2015). *Vandal et al.* (2017) proposed a generalized stacked super-resolution CNN framework for statistical downscaling of climate variables. Super-resolution means a network produces an output image with higher resolution than the input. They argued that a single trained model can downscale spatial heterogeneous regions, and the DL method showed advantages over others. In a recent self-archived paper, authors have employed an MLP to learn fine-resolution dynamics such as convective heating, moistening and cloud-radiative transfer feedbacks from high-resolution simulations, to replace existing multiple parameterization schemes (Gentine et al., 2018). Moreover, the climate modeling community is putting together large datasets to enable big data deep learning on large scales (Racah et al., 2017).

In an interesting application, authors have employed GAN to model urban expansion, which is relevant to water consumption (Albert et al., 2018). Trained on images of 30,000 cities, GAN was able to reproduce realistic concentrations and spreads of urban masses in the absence of externally-imposed constraints, e.g., rules that say cities cannot be built on water. GAN learned these rules by itself.

### *3.1.3. In hydrology*

Compared to some other disciplines, hydrology has not witnessed the wide use of deep learning. Application of DL have been few, especially in a big data setting, and the list of papers reviewed in this sub-section is exhaustive to the author's best compilation effort. There are three types of DL applications in hydrology so far: (i) extracting hydrometeorological and hydrologic information from images; (ii) dynamic modeling of hydrologic variables or data collected by sensor networks; (iii) learning and generating complex data distributions. A common theme reported from these studies is that, unsurprisingly, DL models have surpassed traditional statistical methods. However, there has not been any effort to interpret the DL models.

In the first category, DL has shown strong promise of success for several hydrologic variables, the main one being precipitation. Previously, while satellite-based precipitation retrieval was very useful, its accuracy was still insufficient (Boushaki et al., 2009). As discussed previously in Section 2.2, *Tao et al.* (2016) utilized an SDAE with four layers and 1,000 hidden nodes to extract precipitation from 15 x 15 pixels$^2$ satellite cloud images. They achieved 33%-43% corrections on false alarm pixels compared to PERSIANN-CCS. The authors further improved the results by including water vapor as another channel (Tao et al., 2017). Previously, water vapor was not included in PERSIANN-CCS. To include it in an earlier system like PERSIANN-CCS, significant expertise would have been needed for deriving new features. However, with DL, the authors did not have to specify any formula to realize gains in performance. One could also surmise that a modified CNN will perform even better than SDAE, as CNN is better at capturing spatial texture. In another study, DBN was used to predict field-measured soil moisture, with observed land surface temperature (LST) and leaf area index (LAI) among the inputs from one experiment site (Song et al., 2016). Recall that DBN is an earlier-generation DL method that does not automatically exploit time-dependencies, which is perhaps why real-time data are needed in the inputs. Overall, for information retrieval applications, the use of DL can deliver the most tangible outcomes.

Within the second category, a distinction can be made depending on whether big data are involved. While one of DL's main strengths is to be able to digest big data, it turned out the network architecture can also be trained with data from a single or a few sites. *Bai et al.* (2016) applied non-recurrent DBN to predict single-site time series of inflow to the Three Gorges reservoir. They obtained a good match with observed inflows and concluded that DL was able to engineer multi-scale features. It's worth noting this work learns from the trend, period, and random elements that were extracted from historical inflow time series using Fourier transforms, rather than directly from raw data. Since the sole information used was historical time series, this method should be applied to cases with strong and stable annual cycles and where historical data are sufficient. In an Irish study, a CNN was trained to predict urban water flow given weather forcings and runoff simulated by a hydrologic model as inputs (Assem et al., 2017). This problem would otherwise be difficult to model. Even if one had detailed information about the city's drainage piping system, it could be very time-consuming and error-prone to model. They found CNN to effectively capture urban outflow rate and level in the downstream river. *Zhang et al.*, (2018) built different neural network models, including LSTM and gated recurrent network (GRU, a simpler version of time series DL model), to predict the water level of the Combined Sewage Outflow (CSO) structure. They showed advantages of LSTM over simpler methods. It seems GRU is slightly superior in their case perhaps due to small dataset size. However, the training of these networks with data from one or a few sites would limit the applicability of the network to the training sites. For studies where the dataset size is small, comparisons are necessary to bring out the necessity of DL.

FSKY17 were the first to exploit LSTM's ability to build dynamical hydrologic models with forcings and memory. Their application is relevant to long-term hydrologic hindcast or forecast. They trained LSTM to reproduce surface soil moisture product from the Soil Moisture Active Passive (SMAP) mission (Entekhabi, 2010) over the continental United States (CONUS) using forcing data and outputs from land surface models. The motivation for this work is a data-driven model that faithfully reproduces SMAP data, so that SMAP data can be related to other historical events. In addition, via their approach, the independent information from SMAP sensors can be combined with other land surface models for future projections. However, earlier data-driven methods cannot achieve this goal because they lack the ability to model multi-scale time-dependence while at the same time considering the influences of landcover and soils. In this case, there are thousands of training instances and they built a continental-scale model that benefits from integrating a diversity of data found over CONUS. FSKY17 showed that with 2 years of data, LSTM successfully reproduces SMAP-observed soil moisture dynamics with high fidelity (Figure 4). The deep network is more robust and accurate than traditional statistical methods including regularized linear regression, auto-regressive functions, and a simple feedforward neural network. Their model can be migrated spatially to generate predictions in regions without observations. The prediction results are acceptable as long as the training dataset covers a representative subset of the data. Moreover, in a proof-of-concept test, they found decadal-scale hindcast of simulated soil moisture was as good as annual-scale hindcast (Figure 4a-c). FSKY17 suggests DL is a strong dynamical modeler to unify all available data as opposed to requiring parameter regionalization. The DL model can find commonality among them and identify differences. Another study, currently under open discussion, exploited LSTM in rainfall-runoff modeling (Kratzert et al., 2018). In this work, authors trained LSTM models to predict streamflows for hundreds of basins over CONUS and compared the results to the Sacramento Soil Moisture Accounting Model. They trained models for a single basin or for a group of basins in a region. They did not provide differentiating physical factors for these models, and instead relied on model regionalization to represent spatial heterogeneity. In this sense, LSTM cannot learn how different physical factors play roles in the hydrologic dynamics. Nevertheless, their results suggest that single-basin data was not sufficient to constrain the DL model, as the performance was not as good as the Sacramento soil moisture accounting model, but after fine tuning, DL showed advantages.

In the third category, DL and GAN demonstrated their strengths for the problem of geological media simulation and inversion, which helps the modeling of pollutant transport in groundwater. In this application, the goal is to generate subsurface structures that resemble observed spatial patterns from field samples of porous media. Some traditional methods assume multi-Gaussian distributions and suitable variograms, while some others use multiple-point statistical simulations to exploit higher-order statistics found in data. However, they can be both slow and incomplete for the exploration of the posterior density. In particular, similar to what motivated GAN, these earlier methods tend to be concentrated near a single global minimum and are thus not ideal for multi-modal data distributions. The outcome is unrealistic-looking simulations. Laloy et al. (2017) first attempted to use variation autoencoder (Kingma & Welling, 2013) to extract a low-dimensional representation of the training image, and then produced realizations from it. Recall that autoencoders can store condensed essential features of data (Section 2.3.1). Then, the same authors switched to a spatial GAN to capture the complex spatial data distributions (Laloy et al., 2018). The GAN was notably faster at exploring the posterior model distribution and, because it can capture multi-modal data, it was able to generate images that visually resemble the training images much better. The GAN was trained with much less data than needed for the variational autoencoder. This study is potentially disruptive in that it opens up a more realistic and more uniform road toward geologic inversion and uncertainty quantification, requiring less human interventions.

### 3.2. The budding area of AI interpretation and its application in sciences

The biggest criticism against scientific use of DL networks has perhaps been that they are black boxes that do no help us improve our understanding of the system. When we do not understand the inner workings of a tool, we tend to have lower confidence in its use. However, as DL research progresses, it becomes increasingly possible to use DL as a knowledge discovery tool. Deep networks are not black but gray boxes that can be probed, interrogated and visualized to reveal insights on what is learned by the network. A vibrant sub-field that focuses on DL interpretation and understanding, dubbed by some researchers as *"AI neuroscience"* (Voosen, 2017), has been born. This sub-field is rapidly accumulating a voluminous literature. Systematic and theoretically well-founded methods have begun to emerge. By seeing and understanding the reason why networks make a decision, users of their predictions can become more confident. On the other hand, the gray-box nature of deep networks may in fact be an advantage for scientific research: because they are not bound to humans' pre-conceived notions about how systems function, they might present an opportunity to correct our errors or identify potentially useful linkages that we have not yet noticed.

Most of these AI interrogative procedures conduct posterior analysis, i.e., they seek to obtain insights after the network has been trained. As a summary, interrogative methods originated from the computer science literature can be classified as: (i) backpropagation of decisions or learned features: identifying the part of inputs (or features of input) that are responsible for network decisions or hidden layer representations; (ii) reduced-order approximation: construct simpler, interpretable models to reproduce the response from the full network; (iii) visualization: building prototypes or plotting response functions. In the following subsections, a small, representative list of such studies is covered to introduce the technical ideas. After these methods are described (Sections 3.2.1 to 3.2.3), I review scientific DL interrogative studies in Section 3.2.4.

### *3.2.1. Local approximation by interpretable reduced-order models*

One simple and uniform way of interpreting DL networks is to transfer their knowledge to interpretable models, e.g., regularized linear regression that has only a few predictors. Because these reduced-order models cannot fully describe the complex representations in the deep network, the method is only applied to a region in the input space surrounding the instance to be explained (thus this method is "local"). *Ribeiro et al.* (2016) proposed an explanation system that transfers knowledge from a CNN to a linear regression model. They first broke an input instance down to interpretable components, e.g., patches of an image. Then they perturbed around an instance and selected other instances that are close to the first one, in the feature space. With these samples and their class annotations from the deep network, they then build a lasso-regression model to find approximate outputs from the deep network. In the end, they find patches that explain concepts produced by CNN.

This general concept is also applicable to hydrologic problems. For example, if we have trained a DL network to predict soil moisture given meteorological forcings, we can use this deep network to systematically generate outputs for synthetic cases. The outputs from the network are then used to train a simple model like linear regression, which also predicts soil moisture based on physiographic attributes like soil, slope and landcover. Through this simpler model, we can then understand which physiographic factors have contributed to different response behaviors.

### *3.2.2. Attributing decisions and features to input features*

An "explanation" of model choice can be made by attributing the network's decisions or features engineered by hidden layers to a certain fraction of inputs. If these methods were migrated to hydrology, they might, for example, help visualize what cloud patterns or which cloud patch have contributed to the accurate estimation of precipitation extremes.

#### *I. Backpropagation of decisions*

When AI is trained to choose a word that describes the theme of an image, we can understand its decision by looking at the region on the image that is responsible for this choice. For example, if the algorithm tags as image as having "boat" as a theme, methods exist to find the patch of pixels that most decisively distinguish it as a boat as compared to other possible themes (Bach et al., 2015; Montavon et al., 2017).

*Samek et al.* (2017) developed a simple scheme for CNNs to backpropagate a model decision from the output to the inputs, layer by layer. In this scheme, the network first runs through a forward pass, when the activations for each neuron are recorded. Recall that the activations are weighted and combined to become the input to the neurons on the next level. By recording the activations, we have a way to backtrack which source neurons have contributed to the activation of a neuron on the network, and by how much. Then, the activation at the output layer is backpropagated throughout the network using the recorded contributions. In the end, the procedure generates a heatmap that highlights the importance of each pixel on the input image with respect to the decision of the network (Figure 5).

The decision backpropagation schemes are most relevant to CNNs, because CNNs maintain spatial locality information even at intermediate hidden layers, as explained earlier. It will be more difficult for recurrent neural networks, which are discussed in Section 3.2.3.

*II. Reverse-engineering representations stored in hidden cells*.

Recall that when the DL model is making an inference, inputs propagate from the input layer through the hidden layer to the output layer and the activations from the hidden layers are problem-relevant features engineered by the network (explained in Section 2.3.1). We can gain insights by "reverse-engineering" these features and examining which inputs are used in creating certain engineered features, e.g., edges, color, texture, object parts, etc. The earliest of such work is a deconvolutional network which runs a CNN in reverse order to disaggregate extracted features to the input space (Zeiler & Fergus, 2014). The reverse engineering consists of applying inverse operations to the network calculations. However, recall that there are some non-invertible operations in a CNN: during a forward CNN run, multiple pixels are convolved by a filter, and then the network performs a subsampling, e.g., max-pooling (Section 2.3.2). Therefore, *Zeiler & Fergus* (2014) recorded the position of the maximum value in the lower layer. During their deconvolution, the cell with the recorded location gets assigned the maximum value and other cells get zero. Although simple, the method was able to highlight the pixels responsible for constructing certain features.

A parallel idea is to reconstruct the image using only information stored in the hidden layers (Mahendran & Vedaldi, 2015). The proposed reconstruction (or inversion) procedure solves minimization problems to synthesize input images whose hidden-layer representations best matches the ones to be reconstructed. Then, these generated input images are called the reconstructed images or inversions of the hidden-layer representations. By viewing and comparing these reconstructed images, one can draw insights about what is common and invariant across different instances of the same class. This inversion process is unsurprisingly non-unique, as the same abstract information can be inverted into many different images. As a result, the reconstructed images may also be difficult to understand. The authors' solution was to use a new input image as a "guide" for the inversion, which helps us more concretely understand what transformation has been achieved by hidden layers. The problem is formulated as an activation maximization problem with a new input image as a regularization term: given the class, a new regularization instance, and a hidden-cell representation to be inverted, what is an image that maximizes the output layer's activation value for that class? Gradient ascent was used to solve the maximization problem. Through this approach, *Mahendran and Vedaldi* (2015) showed that some network layers extract photographically accurate information, with different degrees of geometric and photometric invariance, i.e., some variations in these attributes do not influence the network's decision about the class of an image, much like humans can recognize the image of a building even if the picture is slightly distorted.

### *3.2.3. Visualization tools*

It has been increasingly recognized that the weights and activations of neural networks offer clues as to how deep networks function and what they have learned. Therefore, many visualization tools have been created to help users pry into this gray box. Visualization techniques could be the first choice for interpreting DL models created for hydrology, and they are naturally related to correlation-based analysis (Section 3.2.4).

*I. Building prototypes through activation maximization*.

"Interpretation" can be gained by synthesizing *prototypes*: a prototype is an input pattern that is the "most typical" or "most preferred" for a class. For example, for a network that can recognize dog and cat images, synthesizing the prototype instances amounts to answering "what do the most typical/standard dogs and cats look like?" By examining these prototypes, we can either diagnose errors in the model or study the key differences between classes. Borrowing ideas from neuroscientists, the prototype problem was formulated as a maximization problem for the activations at the output layer (Simonyan et al., 2013). The idea is that if an input pattern maximizes the output activation for a class, it can be considered the most standard pattern for that class. Later, some other works extended this framework and added a regularization term which is another input image, called the "expert" (Mahendran & Vedaldi, 2015; Nguyen et al., 2016). Because there are a vast number of input patterns that can generate close-to-maximum activations, the regularization term helps to constrain the maximization such that it is hinged to a real-world input instance, the expert. Metaphorically, the essential features of the class are transplanted to the expert to make the prototype look more realistic. Finally, *Yosinski et al.* (2015) presented an interactive software that integrates the visualization of activations, deconvolved images (Zeiler & Fergus, 2014), and prototype images generated by activation maximization. Using this tool, they demonstrated that feature representations on some CNN layers are local. For example, there are layers in the network that track the positions of faces, text, and flowers, etc., even though the network was not trained with the objective to recognize faces. It means that the network implicitly learned that these objects represent information useful for later classification decisions.

The prototype approach seems quite alien to hydrology. However, if we build DL models to identify plant diseases that result from drought, a prototype might help humans understand the most typical patterns fond on an image that contains infected plants, or the most typical plant and soil moisture conditions that lead to diseases.

*II. Visualization tools for RNNs*

The visualization of RNNs is more complicated than CNNs, as an input may arouse a response from every neuron, and a hidden neuron may be highly responsive to a number of sequences, forming many-to-many relationships (Ming et al., 2017). The visualization tools detailed here can be highly relevant to dynamical models such as ones built in FSKY17. For example, one could examine what kinds of forcing inputs, e.g., increase in vegetation leaf area index, could have triggered certain responses, e.g., the faster-than-normal decline in groundwater levels.

*Karpathy et al.* (2015) were the first to visualize neuron activation functions from a trained LSTM language model and showed how their values respond to text inputs. Their visualization method revealed that there are interpretable neurons: some neurons act as a length counter (on the plot, they increase in color depth as a sentence gets longer, and when it is ended with a period); some respond to opening and closing of quotations, parentheses or punctuations (on the plot, their color turns on inside quotes or parentheses). The neurons are likely to alter other calculations and allow LSTM to "understand" abstract concepts like quotes, which modify the meaning of texts and retain long-range information. LSTM has a large number of neurons. *Karpathy et al.'s* work has mainly focused on how the network internally represents certain functionalities and concepts. This type of analysis is mainly useful from methodological, performance diagnosis, and network design points of view.

*Strobelt et al.* (2016) developed a more generic visual analysis tool for LSTM-language models. The tool, based on parallel-coordinate plots, enables users to select a time window where the phenomenon of interest occurs. Users can then search for hidden states that have high activations within this period but low activation in some other specified periods. Because the selected hidden states can relate to multiple words, the users can then match their input range with other words that could also activate the same highlight hidden states, and by doing so, they can interpret what common patterns have been learned between them. Although this tool was originally designed for language models, it has been adopted to analyze business process intelligence and genomic data. It can be similarly useful for DL in water sciences.

The above studies examined how some neurons respond to certain inputs. However, a neuron's responses can vary based on different contexts. To give an example in hydrology: the same amount of rainfall will produce different runoff amounts when modulated by different antecedent soil moisture. Considering these complications, *Ming et al.* (2017) designed a more advanced visualization system composed of three parts: (1) they calculated expected memory neuron responses to each input word, which is the average of responses to all occurrences of that word from a training database; (2) they conducted co-clustering between neuron responses and inputs. Co-clustering is a technique to consider two dimensions while doing clustering (Hartigan, 1972; Pontes et al., 2015). It allows one instance to appear in multiple clusters; (3) they designed a sequence visualization technique to examine information flows across words. Through these steps, one can observe how a word plays a different role when used in different contexts. While technique (3) is specific to language modeling, type (1) and (2) can be helpful in water-related dynamical modeling because they can reveal possible causal relationships.

### *3.2.4. Interpretative DL studies in sciences*

This section reviews interpretive DL work in sciences in an effort to inspire such work in water. Taking cues from the interpretive work in AI, scientists have started using deep networks as a knowledge extraction tool. However, domain scientists have been creative and have proposed different ways to gain knowledge from DL models. As a summary, scientists have mostly explored (i) direct visualization of cell activation patterns, (ii) correlation-based analysis between cell activations and input features; (iii) domain-specific network design for explaining emergent patterns, and (iv) relevance backpropagation techniques. They have also utilized physics of the problem to expose features learned by the network.

In brain imaging, researchers developed a system that calculates the Pearson's correlations between neuron activations and inputs signals separated into different frequencies (Schirrmeister, Springenberg, et al., 2017). This analysis differs from input-observation correlations. The correlation between inputs and neuron activations reflects *what the model has learned* as important sources of information. Sometimes input-observation correlations are not reliable due to various reasons such as interfering factors, noise, or discontinuous observations (e.g., pathologic vs. normal). Input-activation correlation alleviates these problems. The correlations reflect to which frequency of the input signal the trained network is most sensitive. Via this approach, they showed that the network has learned to apply operations akin to frequency separation when detecting pathology (Schirrmeister, Gemein, et al., 2017). Here the lesson is that we can subset the inputs in different ways and look for high correlations between input subsets and activations. It then informs us what parts of the inputs are the most important. For example, if we hypothetically are to build DL models for human water consumptive demand, we can examine which type of time series signal, e.g., temperature or short-term precipitation, is the most influential for generating changes in water demand, *per the trained model*.

Back to high energy physics, *de Oliveira et al.* (2016) visualized the convolutional layer activations to understand what unique information from jet images was learned by the network to distinguish different particles. They calculated the correlation coefficient between neuron activations and pixels on the image, which revealed the locations with the most discriminating information on the image. The most creative part of their work is that they devised a transformation scheme to remove all discriminating features predicted by known physical laws, and by doing so, they measured what CNN has learned beyond standard physics features. They showed that CNN has uniquely learned features beyond those predicted by physical laws, and located where these features on a phase space plot. They discovered that some information utilized by the CNN was previously unknown to physicists. *Schwartzman et al.* (2016) calculated the correlation coefficient between jet image intensity and activations of hidden states in a trained CNN network. These correlations highlight the pixels on the images with the largest discrimination power. Via this visualization, they found particular radiation patterns that are not exploited in present state-of-the-art physics models. The lesson here is that we can use physically-based mass-conservative hydrologic models to back-out signals that we already know and examine the residuals.

In diagnosing with X-ray images, *Kumar and Menkovski* (2016) employed the deconvolution network approach introduced earlier (Zeiler & Fergus, 2014) (Section 3.2.2) to highlight pixels on images that are most critically responsible for anatomy classifications. They showed that the trained network relied on similar medical landmarks for the classification as human experts, even though the training targets do not explicitly prescribe these features. *Ziletti et al.* (2017) used the same approach to learn about crystal structures that are used by the network to classify crystals. They also found the network identified similar structures of interest as human experts.

Scientists have shown how deep networks can be designed as physics emulators for the phenomena of interest. Trained networks of this kind have the potential to capture emergent behaviors of the system and provide explanations. For example, *Schütt et al.* (2017) developed a deep network with neurons that structurally mimic atoms in the molecule and their interactions. This carefully designed network acquired the high predictive power of molecular properties, with the added advantage that the trained network weights can be interpreted. They showed the network grasped chemical concepts, even though it is solely trained to predict total energy of molecules. In a more recent high-profile neuroscience study, Banino et al. (2018) trained an LSTM connected to a 2D linear layer to predict mammalian navigation. They found that the activations of that 2D linear layer automatically exhibited grid-like spatial response patterns that resemble those cells found in the mammalian cortex. Previously, via brain activity mapping, a Nobel-winning study (Nobelförsamligen, 2014) found that grid-like cells in the brain constitute a positioning system. However, the theory that grid-like cells support computational tasks to achieve navigation goals has not be proven. In Banino et al.'s research, because the AI agent was specifically trained only to perform navigation tasks, it provided strong support for the hypothesis.

The two examples above can be inspiring, as their use of elemental neurons resembles land gridcells in a gridded hydrologic model. If migrated to hydrology, it suggests the elemental hydrologic functioning of an area on land could potentially be derived from data, which could be compared to other process-based understanding. Hydrologists have long paid attention to emergent system behaviors such as river network structure, hydraulic geometry, soil catena, and vegetation patterns (Wagener et al., 2010), but the AI research work mentioned above differs substantially in principle: hydrologists have taken a "Darwinian approach," where they observe emergent behaviors or phenomenological laws. These laws were then used as targets to elicit proposals of organizing principles; or, they have taken a "Newtonian approach," where they put together numerical models, starting from fundamental physical laws, to study whether a given behavior emerges (given a principle, what outcome could have emerged?). In the AI research mentioned above, however, researchers started with some elementary neuron functions and constraints about their connectivity. They trained the networks with observational data in an attempt to constrain the functioning of neurons, and then they examined the emergent properties and functions of the trained network at intermediate layers (given observed outcomes, what fundamental structure could have induced it and what emergent properties does this structure exhibit?). Therefore, this approach can perhaps be called "inverse emergence." While it is not entirely unlike the calibration of conceptual hydrologic models, DL conditions massive amounts of units at the elementary unit level with big data and few assumptions. Model calibration can only condition the parameters but backpropagation also condition the processes: DL networks can be trained to approximate very complex functions and are not constrained to those explicitly prescribed. Hydrologic model calibration cannot modify the mathematical formulations. The interpretive DL approach is thus more similar to SUMMA (Clark, Nijssen, et al., 2015), but it works on a more fundamental level and at a larger scale.

## 4. Tackling water resources challenges with the help of DL

In this section I present some ideas about research applications where DL can help tackle water science challenges presented in the Introduction. I also discuss some potential challenges and possible solutions. First, DL can be water scientists' operational approach to modeling interdisciplinary processes for which mathematical formulations are not well defined but sufficient data exist, especially those related to human dynamics. Some potential applications include modeling land management decisions in response to water constraint or flooding risks, irrigation and consumptive water demands, water saving strategies, ecosystem responses and interactions with landscapes, and urban water flows. For these problems, a process-based model could involve too many inter-dependent processes and parameters without clear ways of obtaining reliable values. The mathematical structure and boundary conditions of these problems and the relationships between model-internal variables could also be too complex and intertwined to fully resolve. For one example, the co-evolution of biome, regolith, climate and terrain may have major controls on catchment hydrologic responses (Troch et al., 2015). Such co-evolution may manifest itself in landscape patterns but its representation is missing in most large-scale land surface models. Myriad interactions exist between these different variables but reliable quantitative theories exposing individual relationships between them remain elusive. For another example, human decisions related to water consumption are affected by their memories of recent extreme events, yet these influences may depend on climate, human development levels and political contexts, etc. (Elshafei et al., 2014). The quantification of these relationships are poor and highly challenging. Ultimately, a better predictive performance may be attained by a holistic and flexible DL model that capture system-wide response patterns. In addition, with the ability to model temporal dependence, DL can readily learn and approximate the effects of collective memory (Di Baldassarre et al., 2013; Viglione et al., 2014) and pendulum swings (Kandasamy et al., 2014) observed in human-water systems, should relevant data exist. It has been shown in health studies that a DL model can predict, with 90% accuracy, whether a person would increase or decrease exercises given their situation in the previous week (Phan et al., 2016). It is reasonable to expect that DL can extract or model collective human behaviors related to water. Experiences from HEP, biology, chemistry, and remote sensing showed that DL can often engineer features on par with or better than human experts (Section 3.1).

DL will prove to be the vital tool to harness the power of big data amidst the emergence of new data sources. With relatively little human intervention, DL can automatically turn raw data into readily useful information, alleviating the stress from the deluge of data. DL can help generate fresh new datasets. Going thought its scientific applications in physical sciences and geosciences, we note that mature DL methods like CNN and LSTM are readily available for the extraction of abstract, high-level information or metrics from images or sequential data. For water sciences, from remote sensing images, DL should be able to identify hydrologic and ecosystem states such as flood inundation, water levels, irrigation amounts, precipitation, vegetation stress, and other observable events. While previously algorithms have been developed for some of these variables, the advantage of DL is improved accuracy (as it extracts high-order features such as textures), more uniform training procedures (without the need for expert-guided case-by-case corrections), and the possibility for transfer learning (making use of massive data from generic sources). Monitoring data collected from smart water infrastructure can be mined to model otherwise unwieldy problems for improved controls of safety, resilience, and quality. Optimal controls of smart water infrastructure can also be implemented via DL. Sensor data can also help farmers gauge crop conditions and DL can perhaps better predict crop productivity. There are many other sorts of big water data, some arising from unconventional sensors, to be harnessed.

DL-based data-driven models can measure whether a new source of information could help model a phenomenon of interest. Scientists may even use DL to gauge the possibility that two seemingly unrelated variables are in fact related, to discover unrecognized linkages. Along this line of thinking, DL can help inexpensively uncover relationships between landcover, climate, soil, and geomorphology. For an example, DL can detect if Karst geology or bedrock outcrops significantly alter hydrologic responses, and what those influences could be. While data mining methods have long existed, based on experiences from HEP and chemistry, DL offers the unique expressive power to integrate big data, engineer features, and work with lower statistical thresholds.

In both of the aspects mentioned above, the “off-the-shelf” DL algorithms or some existing variants could suffice, and the most significant challenge may be to obtain the needed supervising datasets. For standard AI work, the industry has relied on human labelers, e.g., Mechanical Turk to create large volumes of labeled data. For water scientific applications, in most scenarios, there already are some existing observations and methods of identification. Borrowing experiences from other domains, to reduce the amount of workload, research work should carefully examine possibilities for transfer learning, where they can start with publicly available networks such as GoogLeNet (Szegedy et al., 2015). Then the researchers need to modify the structure of the output nodes to fine-tune the network to their applications. How to organically integrate these available resources presents some research opportunities apart from the main research questions.

DL may offer a fresh opportunity at hydrologic modelers’ greatest “nemeses,” including scaling and equifinality. DL has already shown promise in facilitating climate model downscaling (Section 3.1.2). By learning from high-resolution simulations, DL can help capture the effects of resolution on outcomes. Research work is further needed to customize networks for hydrologic modeling, which differ greatly from climate modeling in that it involves multi-scale processes constrained by highly heterogeneous physical geology, topography, vegetation, and humans, etc. As these heterogeneities are not the standard for DL networks, research is needed to efficiently incorporate them into the network structures. Because DL inference can execute rapidly on either on CPU or specialized hardware like GPU, it can greatly reduce the computational burden of high-resolution simulations. Given its ability to create a uniform model that digests all data, DL can help predict effective model parameters, e.g., runoff coefficient or variable infiltration curve parameters, from raw data. We can use DL to uniformly parameterize large-scale models, without regionalized customizations. This kind of global (or continental-scale) parameterization is constrained by large-scale and diverse datasets and could thus be less prone to overfitting to regional patterns. Because regionalization is avoided, there are far more data to train the global-scope network. Therefore, it may also help address the equifinality challenge as well as the regionalized parameter challenge.

At first glance, the GAN may seem like an alien and irrelevant concept to water scientists, but in fact, hydrologists have long used stochastic generators for variables ranging from weather to the hydraulic conductivity field. Given experiences from those described in Sections 3.1.2 and 3.1.3, GANs are especially suited as generators for weather, vegetation, and soil features, because they not only learn relationship with covariates but also respect the spatial structure to high orders. It also does not require an extensive mathematical specification for the error model. There are significant research opportunities in terms of using GANs that respect observed constraints. As discussed earlier, realizations generated by GANs are typically more realistic-looking and sharper to the human eye, even if the difference may not manifest in statistical metrics. In addition, hydrologists have long pursued the idea of “doing hydrology backward” (Acharya et al., 2017; Kirchner, 2009); that is, inferring the driving forces from the outcomes. There are many other water problems where we can observe consequences but not the forcings, e.g., surface water and groundwater contamination. Realizations from such inversions can aid uncertainty quantification. Hydrologic variables such as soil moisture, precipitation and groundwater residence times exhibits multi-modal spatial distributions (McCallum et al., 2014; Ryu & Famiglietti, 2005). GANs provide a new and invigorating way to systematically explore this inversion.

In addition to helping address these challenges, quantitative changes in predictive accuracy sometimes bring forth qualitative change in our understanding. For example, after observing how a complex chaos (predicated by a dynamical system equation called the Kuramoto-Sivashinsky equation) evolves for a period of time, machine learning can accurately predict chaotic attractors and Lyapunov exponents out to 8 Lyapunov times (8 times further than previous methods), purely in a model-free, purely data-driven fashion (Pathak et al., 2018). Previously data-driven methods were not deemed plausible for complex chaos like this. In hydrology, if we came across better-than-expected predictive accuracy, it may lead us to refresh our understanding of limits.

Echoing the discussion in Section 3.2.4, in terms of explaining emergent behaviors, the backpropagation procedure can directly and efficiently exploit the value of big data in conditioning parameters as well as physical processes, an advantage not shared by hydrologists’ traditional approaches. Therefore, it could offer a novel and constructive way of finding hydrologic principles. It is possible that we can borrow this approach to train networks or models to approximate hydrologic and human-hydrology systems, to shed light on fundamental functioning of its elementary units. One might also design emulator networks that resemble real-world subsystems, but with minimal constraints on formulations, to infer system functions that lead to certain emergent dynamics. This network could be built on an elementary representative unit level, e.g., a land gridcell, or an organizational unit level, e.g., a water consumption community.

## 5. Limitations and potential issues of DL

The training of GANs can be quite challenging (Radford et al., 2016) and can sometimes be unstable: It can be difficult to get the generator and the discriminator to converge; the generative model may collapse and generate very similar results; the discriminator may become too strong, its loss approaching zero too rapidly so that it leaves little gradient information for the generator to learn from, etc. (Creswell et al., 2017). The training of GANs may require tricks and significant experimentation. Computer science expertise should be sought for the application of GANs.

Related to GANs, deep networks can sometimes be fooled when they are under "adversarial attack," that is, small, unperceivable perturbations to inputs sometimes cause large changes in predictions, leading to incorrect outcomes (Szegedy et al., 2013). Adversarial attacks pose a significant security threat to AI applications, especially in computer vision problems. A rapidly rising number of studies focused on how to systematically generate these malicious perturbations, and this sub-area is at the front of AI research (Akhtar & Mian, 2018). While RNNs are not immune to adversarial attacks (Papernot et al., 2016), most of these studies are for classification tasks. Little research has examined the extent of the issue for regression tasks, and it is almost blank for scientific applications. It would be reasonable to surmise that adversarial attacks are a smaller issue for continuous predictions. Some have argued that the existence of adversarial examples is due to model overfitting and nonlinearity, while others argued, surprisingly, that they are due to the linear nature of deep network predictions (Goodfellow et al., 2015). Nevertheless, once we are able to seek out ways of generating these adversarial examples, they can be included in the training instances to train DL models more robustly, so they are resistant to such challenges (Ororbia et al., 2016).

DL models require large sample sizes. While some studies introduced in Section 3.1.3 used data from single sites, they are only exploiting the usefulness of network architecture, and are underutilizing the model's ability to build complex models for large datasets. It is clear from previous experiences that when data are limited, simpler models with stronger assumptions about the structure of the problem are better constrained (Kratzert et al., 2018). DL's advantages increase as data size scales up.

Despite the comparisons between DL and non-deep machine learning, this author would strongly advise against applying DL non-discriminatively. The earlier-generation methods can be highly valuable for their respective problems and situations, especially when there are limited data or relatively homogeneous data. As shown in studies, for small dataset, DL could be at a disadvantage compared to models with stronger structural assumptions. Methods like random forest and Gaussian processes can be powerful and robust for many inference problems and have additional benefits such as uncertainty estimates. Non-deep machine learning methods are also appealing for well-defined problems where there are significant accumulations of expertise or features. In addition, some of these methods, e.g., CART and SVM, are easy to interpret and could be used to approximate full-complexity models in a certain region of the input space (Section 3.2.1).

## 6. Concluding remarks

Deep learning represents a potentially disruptive force for many scientific disciplines. Its ability to engineer features by itself and capture high-dimensional, multi-modal data distributions stand out as its defining strengths, in terms of not only practical benefits (efficiency and not requiring significant expertise) but also uncovering features beyond current knowledge. The clearing of key roadblocks such as multiple-scale representations, overfitting, and efficient computing platforms enabled exponential growth in DL-powered applications. For water sciences, DL could help tackle several major challenges, old or new. Vast opportunities exist for DL to propel advances in water sciences.

Indeed, from the literature we can see that DL serves roughly two functionalities: (A) to build models with more accurate predictions, far-greater processing capability, and reduced demand for human involvement and expertise; and (B) as an exploratory data mining tool to support discoveries that expand present knowledge and capabilities. The astronomy, remote-sensing, and climate studies primarily utilized aspect (A) of DL. In this regard, the supervised learning aims at reducing human effort and improving data discoverability. One can notice emerging aspects of (B) in applications from HEP, computational biology and chemistry because there the "supervision" arises from observations of nature. For example, if DL models make predictions about new binding sequences of proteins, new compounds, or particles that humans do not know yet, it may prompt the reexamination of theories and new experiments.

Synthesizing experiences from many disciplines, if the objective is retrieving information from images or sequential data, the default architecture of CNN and LSTM should be good starting points for research. When data organization is unlike 2D or 3D images, SDAEs are useful elements to have in the network design. When the aim is to uncover "inverse emergence," network architectures need to be designed from the ground up. Multi-task learning, in general, can provide some incremental gains. If transfer learning is applicable, it can generally lead to substantial boosts in performance because it implicitly makes use of massive datasets. Therefore, new water research should investigate these approaches if applicable.

While DL interpretive research appears to be a blank thus far for water resources applications, one easily witnesses the fast growth in the young and energetic field of AI neuroscience. As these studies arise from AI research, they generally set their focus on understanding how deep networks function. On one hand, computer scientists are paying more and more attention to the explaining of network decisions and are developing "standard" methodologies. On the other hand, domain scientists have been creative in devising original methods that suit their particular situations, such as specialized architecture and the use of governing equations to remove known signals. There are extensive research opportunities in terms of incorporating either methodology in understanding network decisions. Many scientific investigations visualize the correlations between hidden activations and inputs. This approach is the most achievable and should be valid as a first-order screening for locating potentially important features. However, as the units in a hidden layer store the representation in a distributed way, only the combined effects of activation cause the correct outputs. Interpreting the effects of particular cell activations in isolation can provide incomplete and inconclusive analysis. Therefore, the holistic interpretive tools, e.g., reduced order models, may better reflect the inner workings of the DL model than correlation-based analysis. However, this argument should not prevent us from using the latter as early exploratory and visualization tools. Relevance backpropagation methods can point to inputs that explain the cause of events, and would typically be more applicable for problems with spatial dimensions and those similar to image recognition, e.g., precipitation retrieval. It will perhaps be difficult to implement for recurrent networks. Developing emulator network structures will help solve fundamental mechanisms leading to emergence, but it will require substantial expertise in computer science to construct viable and efficient models. Therefore, strong collaboration with computer scientists will be needed.

## Appendix A. Some early machine learning methods and their hydrologic applications

This section briefly reviews hydrologic applications of several machine learning algorithms. The concise outline here cannot possibly do justice, and readers are referred to multiple review papers for more details, (e.g., Dawson & Wilby, 2001; Fallah-Mehdipour et al., 2013; Jha & Sahoo, 2015; Raghavendra & Deka, 2014; Tsai et al., 2017; W.-C. Wang et al., 2009; Yaseen et al., 2015).

Earlier-generation machine learning methods such as Support Vector Machine (SVM) (Cortes & Vapnik, 1995; Vapnik, 1992), Classification and Regression Tree (CART) (Breiman et al., 1984), and ANNs have been applied extensively in hydrologic and land surface predictions to moderate success. Among these, Classification and Regression Tree is an effective data-driven strategy to split data points into a hierarchy of bins by thresholds in the predictors to reduce variance within bins (Loh, 2011). CART is not always highly accurate, but it has the advantage of presenting interpretable hypotheses in the form of the hierarchy of decision criteria: by examining the split criteria, one can form hypotheses about major driving forces that result in different outcomes. In hydrology, *Sawicz et al.* (2014) used CART to present a coherent classification for the catchments based on their hydrologic signatures; White et al. (2005) used regression tree to identify the controls of inter-annual vegetation variability, and found that vegetation responses to climate oscillations depend on a variety of topographic attributes such as elevation, slope, aspect, and proximity to groundwater convergence zones; *Fang and Shen* (2017) used CART to form hypotheses about possible physical causes that impact the correlations between streamflow and satellite-measured water storage. The CART criterion led them to hypothesize that groundwater controls the full range of streamflows in the southeastern CONUS *because of* the thick soils, and thin soils along the Appalachian Plateau impose limits on water storage. However, one should not attempt to interpret lower-level splits because they often have few data points and are thus unstable. For this reason, CART is also said to be statistically inefficient (Frosst & Hinton, 2017). To improve predictive power, one can combine the prediction from an ensemble of classification trees, which is called random forest (Ho, 1995; Zhou & Jiang, 2004). *Chaney et al.* (2016) used random forest and environmental co-variates to estimate soil properties at 30-*m* resolution for CONUS. This is a well-suited application of the random forest. However, random forests are usually very hard to interpret.

SVM attempts to create optimal separating hyperplanes in the input space between classes to maximize the margin between classes. A version of SVM, called support vector regression (Drucker et al., 1996), builds regression models using only a subset of the training data and ignores these too far from the model predictions so that the model becomes more robust. SVM is built from elegant statistical theory and hydrologists have compared SVM with simple ANNs, multiple linear regressions, and other methods and found it to be robust (Raghavendra & Deka, 2014). SVM performed comparatively well in predictions of reservoir inflow (Lin et al., 2006), soil moisture retrieval using vegetation indices and precipitation (Ahmad et al., 2010), streamflow (Asefa et al., 2006), evapotranspiration (Tabari et al., 2012), lake level (Khan & Coulibaly, 2006), groundwater level (Yoon et al., 2011), flood stage (Yu et al., 2006), and sediment yield (Çimen, 2008), among others. SVM was also shown to be useful in statistical downscaling of atmospheric forcings (H. Chen et al., 2010; Tripathi et al., 2006). Researchers have coupled SVM with spectral analysis to predict runoff, where spectral analysis decomposes frequency components while SVM helps in generalization performance (Sivapragasam et al., 2001). On the topic of model correction, *Xu and Valocchi* (2015) proposed an uncertainty quantification system, where they built a data-driven model (either random forest or SVM) to capture the bias of a calibrated groundwater model. They then applied the bias correction to separate out the epistemic (bias) and aleatory (variance) part of the uncertainty. Most of these studies built models using data from one or a few sites.

Genetic Programming (GP) (Koza, 1992) is a special data-driven technique that aims at prescribing explicit mathematical formulas. Just as SVM or ANN, GP also models the mapping relationship between inputs and outputs, but it centers around building "programs," which are symbolic formulas that are executable and are represented internally as a graph. Starting from a small number of parent programs, GP evolves many offspring programs via genetic or evolutionary operations such as cross-overs, selections, and mutations, with each offspring program's performance evaluated with respect to the training objective. To avoid generating formulas with large complexity (an issue called bloating), bloat control methods that penalize complexity could be employed (Silva & Costa, 2004). *Savic et al.* (1999) used GP to create a rainfall-runoff model whose formula only relied on rainfall from previous days. Researchers used GP to create models for monthly groundwater levels from several wells, where GP showed good performance but prescribed complex functions of rainfall forcing and groundwater level in past time steps (Kasiviswanathan et al., 2016). When GP was used to predict saturated hydraulic conductivity ($K_s$), it generated a simple, regression-like equation (Fallah-Mehdipour et al., 2013). Another similar study found that $K_s$ is larger when there is a higher sand percentage, but bulk density has a negative impact (Parasuraman et al., 2007). Similar regression-like equations were given for modeling evapotranspiration (Izadifar & Elshorbagy, 2010). However, when GP was used to predict the soil water retention formula, it generated a very complex formula, perhaps as a result of insufficient bloat control (Garg et al., 2014). GP was also used in flood routing (Sivapragasam et al., 2008), and streamflow forecasting (Guven, 2009; Makkeasorn et al., 2008). GP can be computationally intensive, and recently it appears it has not been a focus of machine learning research, especially for complex problems in the big data domain.

A large body of hydrologic literature is devoted to ANNs, attesting to the long history hydrology has had with neural networks. Its applications extend to rainfall-runoff modeling, streamflow, groundwater management, water quality, stream salinity, rainfall estimation, ocean wave heights, and real-time forecasting (Bowden et al., 2005; Dawson & Wilby, 2001; Deo & Naidu, 1998; Govindaraju & Rao, 2000; Gupta et al., 2000; Maier et al., 2010; Maier & Dandy, 1996; Sorooshian et al., 2000). Also see the collection of articles in (Govindaraju & Rao, 2000). ANN is suitable for preparing inputs for models. *Schaap et al.* (2001) trained ANNs to store soil pedo-transfer functions, which are widely used in hydrologic models, e.g., (Clark, Fan, et al., 2015; Fatichi et al., 2016; Ji & Shen, 2018; Maxwell et al., 2014; Shen et al., 2016). *Hsu et al.* (2002), proposed a self-organizing linear output map for hydrologic modeling and analysis. *Abramowitz et al.* (2006) used ANN to predict the error of a land surface model and reduced annual error by 95%. An ANN trained to work at one biome completely corrects the error for another in a large temperature range (Abramowitz et al., 2007). Multiple studies have used ANNs to retrieve variables from satellite images (Aires et al., 2001; Kolassa et al., 2017; Rodriguez-Fernandez et al., 2014; Tapiador et al., 2004). Hydrologists have often not paid enough attention to network geometry and transfer (or activation) functions, but these configurations have important impacts on training speed and generalization impact (Maier & Dandy, 1998). Nevertheless, there seem to be few studies that have attempted to interpret the knowledge extracted in ANNs.

Gaussian Process Regression (GPR), related to Kriging in Geostatistics, relies on modeling a covariance function, or a kernel, which describes a notion of smoothness between near data points (Rasmussen & Williams, 2006). The hyper-parameters of the kernel function can be learned during the fitting (or training) procedure. A nice feature of GPR is that it naturally yields estimates of posterior uncertainty, which may manifest as larger prediction bands for instances or time periods that are far from known data points. In hydrology, researchers have employed GPR to make predictions for streamflow (A. Y. Sun et al., 2014; J. Yang et al., 2018), precipitation (Kleiber et al., 2012), and soil moisture (Andugula et al., 2017), etc. Challenges facing GPR include computational efficiency, expressive power, the use of non-Gaussian likehoods, and the scaling to large datasets. There are now developments in deep Gaussian Process, which is a deep belief network where input variables are mapped to the output through a cascade of hidden layers, where transformations between layers are modeled with Gaussian Processes(Damianou & Lawrence, 2013). In a unique series of studies, *Gong et al.* (2013) proposed a quantitative method, grounded in information theory, to separate out uncertainties due to inputs and model structural error. The main idea is to estimate the best achievable performance (BAP) of *any* model given forcing and benchmark observations, using a model-independent statistical method, in their case the Sparse Gaussian Processes (Snelson & Ghahramani, 2006). *Nearing et al.* (2016) further separated out the uncertainty due to parameters by including models that are trained locally. The information resolvable by the BAP is measured by the mutual information (*MI*, with the same unit as entropy) (Cover & Thomas, 1991) between forcing data and observed diagnostic variables.

## Acknowledgments

Shen was supported by U.S. Department of Energy Office of Science under contract DE-SC0016605, the U.S. National Science Foundation under award EAR-1832294, and the Penn State Institute of CyberScience seed grant. The present work was partially developed within the framework of the Panta Rhei Research Initiative of the International Association of Hydrological Sciences (IAHS). Shen would like to thank Dr. Thomas Burbey from Virginia Tech, Dr. Gordon Grant from US Forest Service, Dr. Phanikumar Mantha from Michigan State and three anonymous reviewers whose suggestions and comments have helped to improve this paper greatly. This review paper is theoretical and does not contain any dataset to be shared.

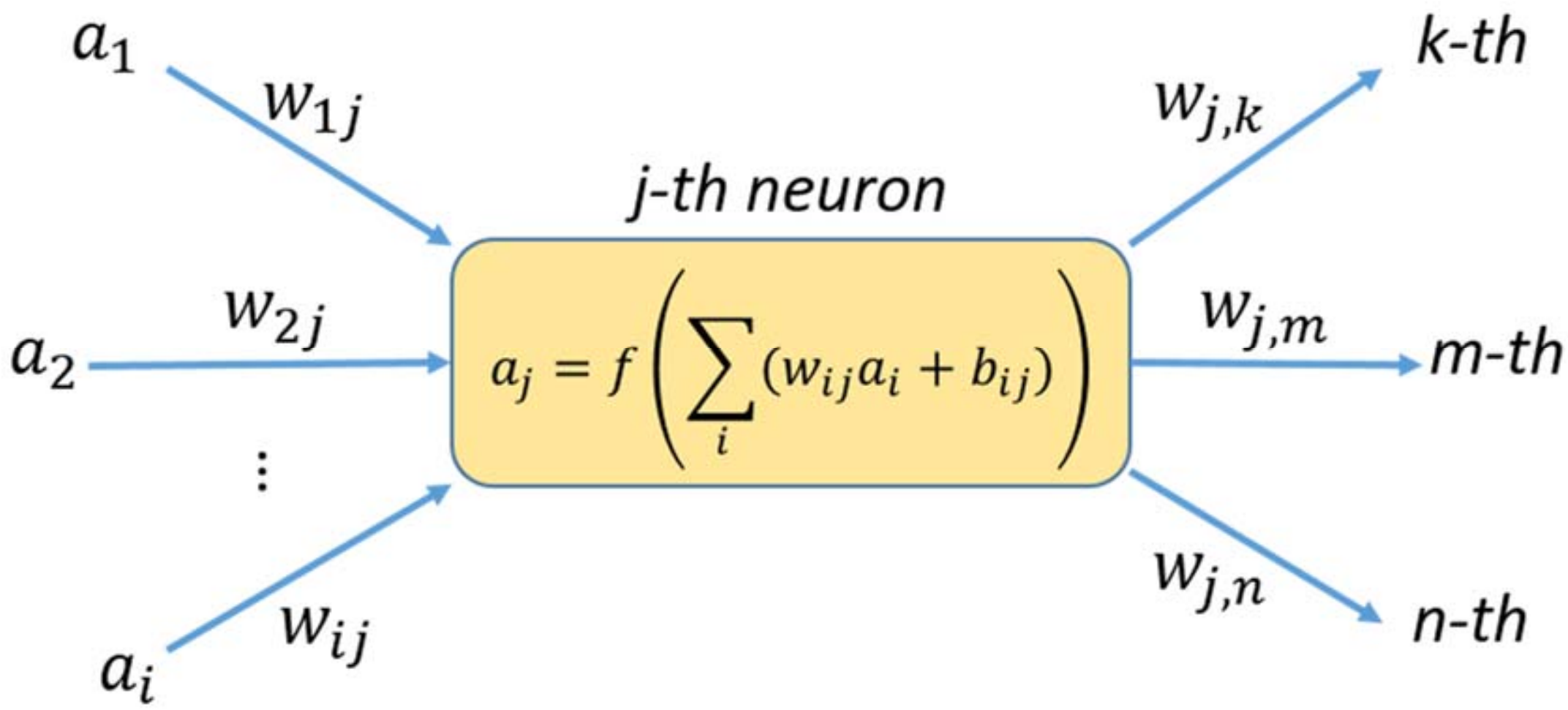

**Figure 1. Illustration of the functioning of a standard neuron in a feedforward network. The *j*-th neuron receives inputs from neurons (1, 2, …, *i*) on the upstream layer multiplied by their respective weights, applies a nonlinear transformation (*f*). The result of the computation is called the activation of this neuron, and is sent to downstream cells. Note this the standard neuron and there could be other designs for more specific purposes. $w_{ij}$ and $b_{ij}$ are the weights and the bias parameters, respectively, and they are adjusted during training.**

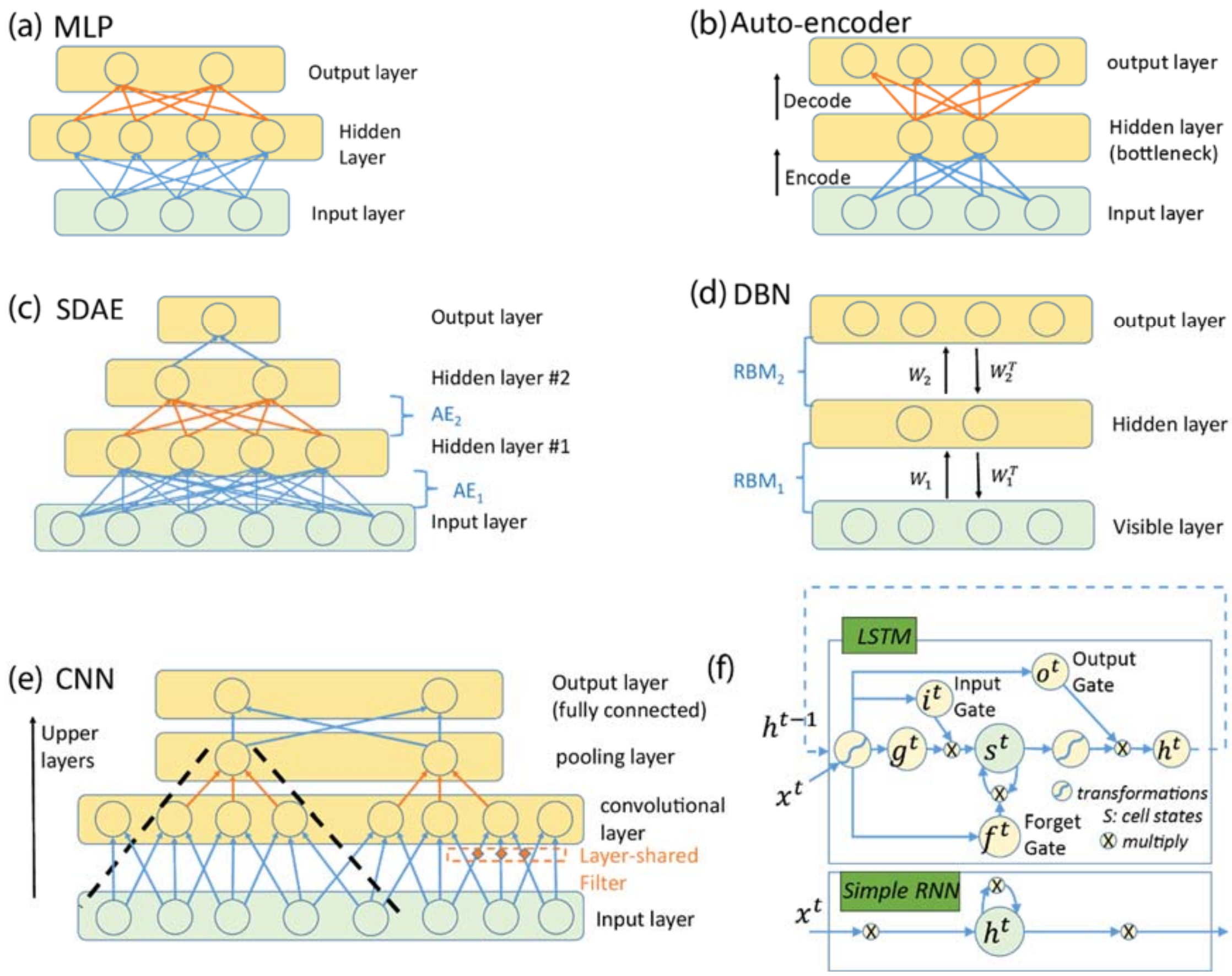

**Figure 2. The architecture of components of several neural network structures. Layers are visualized as 1-dimensional arrays, but they can be 2D or 3D. (a) A "vanilla" multilayer perceptron network (MLP). Here, each circle represents a neuron (explained in Figure 1). The layer in the middle is not connected to input or output and is thus called the "hidden" layer. The layers are "fully connected", meaning that a neuron in a layer is connected to all cells in the adjacent layers. (b) Autoencoder: note the hidden layer has fewer neurons than input and output layers, which have the same number of neurons. This autoencoder is trained to reproduce its inputs through the bottleneck layer in the middle, which is forced to apply information compression; (c) Stacked denoising autoencoder (SDAE), formed by stacking autoencoders: each hidden layer is first trained to reproduce its inputs, then activations from this layer is sent to the next hidden layer, which has fewer neurons, as the "input data" to be reproduced. The layers are fully connected. (d) Deep Belief Net (DBN) stacks layers of Restricted Boltzmann Machines (RBMs) with a specific order in the number of their elements. Each RBM is trained to stochastically reproduce its input distributions after going through a forward and reconstruction step. (e) Convolutional Neural Network (CNN): although it structural resembles SDAE, each coarsening stage (reduction in the number of neurons) forward contains multiple convolutional layers and pooling layers (here only one set is illustrated). The convolutional layer convolves a layer-shared filter with the inputs, while the pooling layer employs a reduction operator, e.g., max, to coarsen the layers. There can be many such coarsening stages. Thus a neuron in the upper-level layers has a "field of view" indicated by the dashed black lines. (f) Comparing Long Short-Term Memory (LSTM) and simple recurrent neural networks (reprinted from with permission). $\otimes$ means multiplication by weights. In this plot alone, activation functions are explicitly shown as transformations. The design of gates allows LSTM to learn when to forget past states, and when to output.**

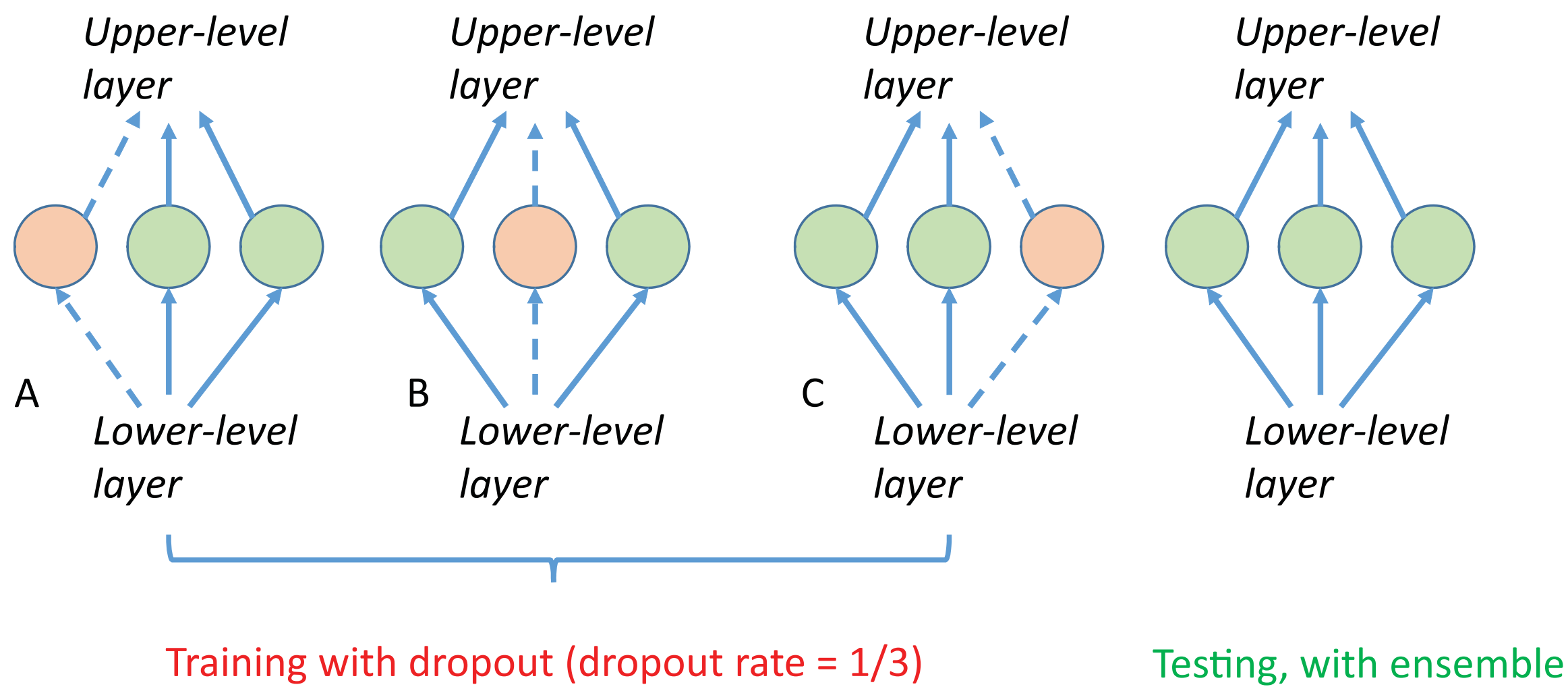

**Figure 3. An illustration of the dropout regularization. During training, some of the connections to neurons in the dropout operand layer are stochastically set to 0 (indicated by orange color and dashed arrows). A, B, C are three realizations of dropout masks and can be regarded as sub-networks. The dropout mask changes of the dropout operator, e.g., for training with each batch of instances. All weights are employed during testing, which is equivalent to the ensemble predictions of the subnetworks.**

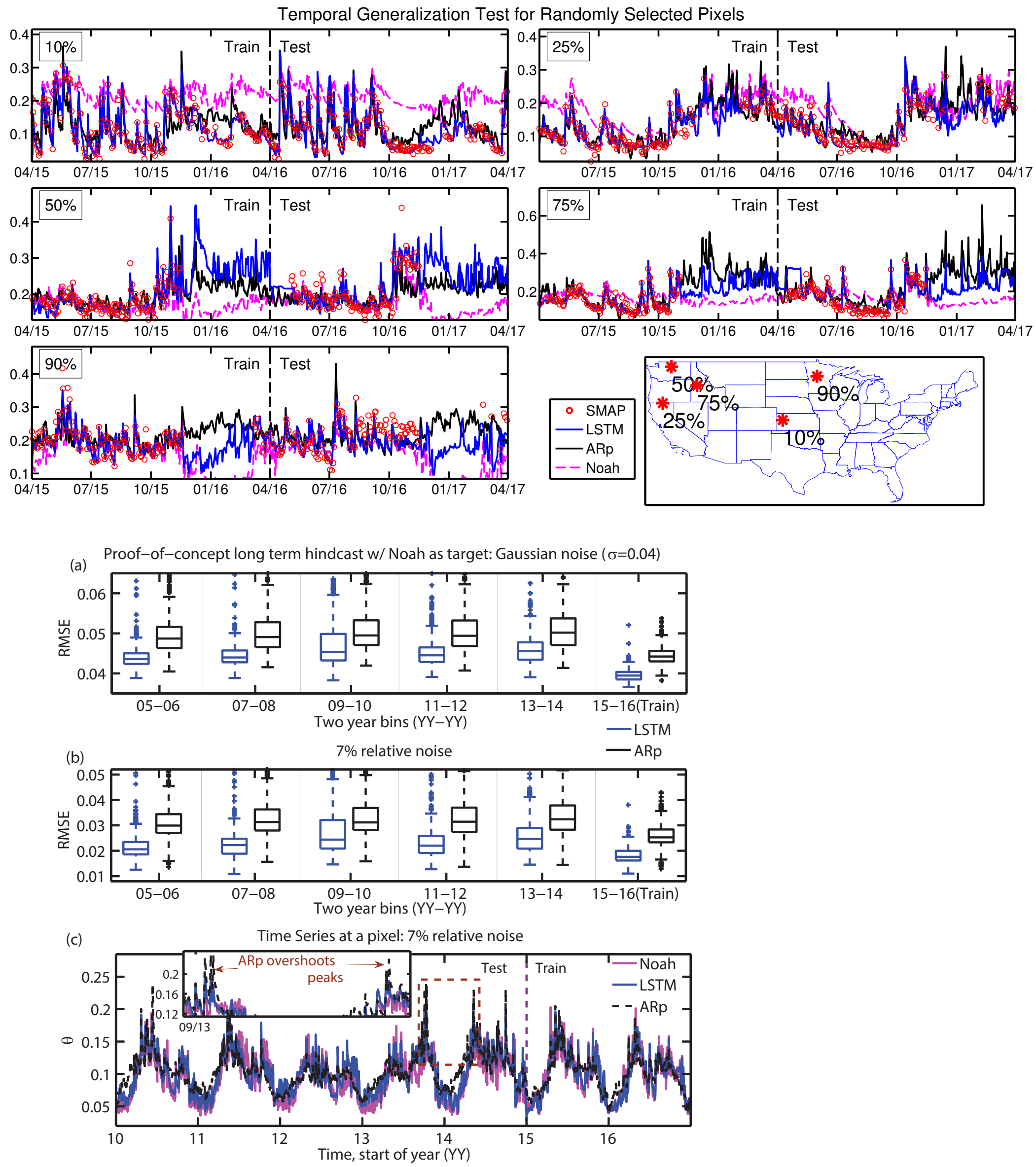

**Figure 4. Upper panels: Comparisons between SMAP observations and soil moisture predicted by LSTM, the land surface model Noah, and Auto-regressive model at five locations. We chose sites around 10th, 25th, 50th, 75th, and 90th percentiles as ranked by the correlation coefficient between LSTM and SMAP; Lower panels (a-c) proof-of-concept long-term hindcasting experiments using noise-contaminated Noah solution as the target. The results indicate long-term hindcasting of soil moisture using LSTM is very promising. Figures were recreated from FSKY17 with permission.**

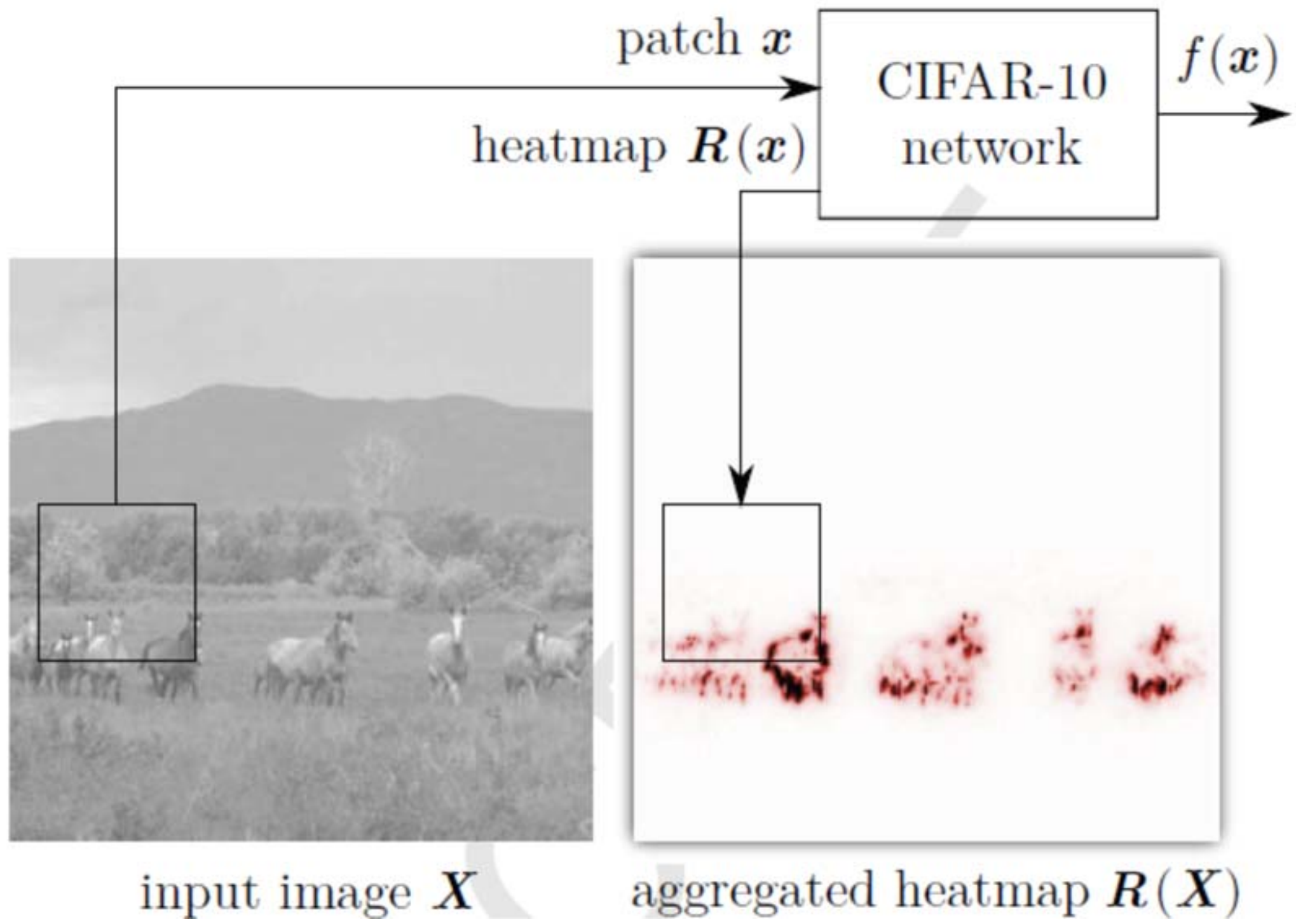

**Figure 5. Using (windowed) layer-wise relevant propagation to identify pixels that contribute to the identification of an image as the class "horse." Reprinted from** (Montavon et al., 2017) **under Creative Commons Attribution License.**